\documentclass[submitting]{nst}
\usepackage{subfigure,dcolumn}
\usepackage[T2A,T1]{fontenc}
\usepackage[russian,english]{babel}
\usepackage{amsmath}
\usepackage{bm}
\usepackage[ruled,vlined]{algorithm2e}
\usepackage{algpseudocode}
\usepackage{tabularx}
\usepackage{graphicx}
\usepackage{subfigure} 
\usepackage{listings}
\usepackage{float}  

\begin{document}

\title{Residual resampling-based physics-informed neural network for neutron diffusion equations}\thanks{Supported by the Science and Technology on Reactor System Design Technology Laboratory (No.LRSDT12023108) and supported in part by the Chongqing Postdoctoral Science Foundation (No.cstc2021jcyj-bsh0252), the National Natural Science Foundation of China(No.12005030), Sichuan Province to unveil the list of marshal industry common technology research projects(23jBGOV0001) and Special Program for Stabilizing Support to Basic Research of National Basic Research Institutes(WDZC-2023-05-03-05)}

\author{Heng Zhang}
\affiliation{Chongqing University of Posts and Telecommunications, Chongqing 400065, China}
\author{Yun-Ling He}
\affiliation{Chongqing University of Posts and Telecommunications, Chongqing 400065, China}
\author{Dong Liu}
\email[Corresponding author, ]{Dong Liu,No. 328, Section 1, Changshun Avenue, Huayang, Shuangliu County, Chengdu, Sichuan, China,13980956120, liudong73@yahoo.com}
\affiliation{Science and Technology on Reactor System Design Technology Laboratory, Nuclear Power Institute of China, Chengdu, 610213, China}
\affiliation{CNNC Engineering Research Center of Nuclear Energy Software and Digital Reactor, Chengdu, 610213, China}
\author{Qin Hang}
\affiliation{Chongqing University of Posts and Telecommunications, Chongqing 400065, China}
\author{He-Min Yao}
\affiliation{Chongqing University of Posts and Telecommunications, Chongqing 400065, China}
\author{Di Xiang}
\affiliation{Science and Technology on Reactor System Design Technology Laboratory, Nuclear Power Institute of China, Chengdu, 610213, China}
\affiliation{CNNC Engineering Research Center of Nuclear Energy Software and Digital Reactor, Chengdu, 610213, China}

\begin{abstract}
The neutron diffusion equation plays a pivotal role in the analysis of nuclear reactors. Nevertheless, employing the Physics-Informed Neural Network (PINN) method for its solution entails certain limitations. Traditional PINN approaches often utilize fully connected network (FCN) architecture, which is susceptible to overfitting, training instability, and gradient vanishing issues as the network depth increases. These challenges result in accuracy bottlenecks in the solution.
In response to these issues, the Residual-based Resample Physics-Informed Neural Network(R$^2$-PINN) is proposed, which proposes an improved PINN architecture that replaces the FCN with a Convolutional Neural Network with a shortcut(S-CNN), incorporating skip connections to facilitate gradient propagation between network layers. Additionally, the incorporation of the Residual Adaptive Resampling (RAR) mechanism dynamically increases sampling points, enhancing the spatial representation capabilities and overall predictive accuracy of the model.
The experimental results illustrate that our approach significantly improves the model's convergence capability, achieving high-precision predictions of physical fields. In comparison to traditional FCN-based PINN methods, R$^2$-PINN effectively overcomes the limitations inherent in current methods, providing more accurate and robust solutions for neutron diffusion equations.
\end{abstract}

\keywords{Neutron diffusion equation, Physics-Informed Neural Network, CNN with shortcut, Residual Adaptive Resampling}

\maketitle

\section{Introduction}\label{sec:1}
Nuclear reactor core analysis is crucial for ensuring safe operation of nuclear reactors. The neutron diffusion equation, which describes neutron movement within a medium, is fundamental to this analysis\cite{he}. Numerical methods are widely applied in many physical scenarios. Methods such as finite difference \cite{ozicsik2017finite} and finite element \cite{younes2010mixed} have been continuously developed and improved. In the reactor domain, many related works can be found. For example, Yasser Mohamed Hamada\cite{hamada2022higher} proposed higher-order compact finite difference schemes for solving neutron diffusion equations. At the same time, Yuk\cite{yuk2020time} utilized the finite element method to solve the time-dependent neutron diffusion equation. X Li\cite{li2024research} designs an algorithm based on the finite volume method to solve multigroup neutron diffusion equations. For the same problem, Matheus Gularte Tavares\cite{tavares2021solution} and K Zhuang\cite{zhuang2021variational} use the Source Iterative Method and variational nodal method to solve, respectively. Additionally, lots of researchers have successfully employed CFD software such as COMSOL\cite{huang2023solution}, ANSYS FLUENT\cite{sidi2023neutronic} and OpenFOAM\cite{ma2021ntkfoam} to tackle neutron diffusion problems.

However, these methods require the discretization of the solution domain, which can be computationally complex and time-consuming when high-precision physics reconstruction is required\cite{dimensioncurse}—also, considering the complex environment in nuclear reactors, which exist in multi-physics fields such as neutron transport and heat transfer. Numerical methods have to simplify and approximate the model in solving the equations, which will introduce a certain amount of solution error. Meanwhile, benefiting from the development of neural networks (NN), there has been increasing interest in machine learning-based approaches\cite{summary}. Training NN makes it possible to predict the physical field faster than traditional numerical methods. Then, engineering prior knowledge should be incorporated into the network to make NN predictions more consistent with physical laws. PINNs, introduced by Raissi\cite{raissi2019physics}, incorporate partial differential equations(PDEs) as constraints during network training by increasing penalties for violating PDE data points. This leads to higher precision and physically consistent solutions. Until now, PINNs have been successfully applied to various scientific computation problems in engineering fields such as fluids\cite{cai2021physics,baochang2023self}, heat transfer\cite{cai2021physics2,YUBo}, flows\cite{mao2020physics,ChengC}, and solid mechanics \cite{bai2023physics,TangsijieF}, yielding significant results. The difference between PINNs and traditional numerical methods are shown in Table~\ref{tab:compare}. In nuclear reactor core analysis, the Monte-Carlo method\cite{MTCL_LIUPeng,MTCL_guo2023development} and CFD\cite{CFD_reactor} can also be used to solve reactor problems. In recent years, Deep neural networks (DNN) have been more frequently used in the problem of reactor core\cite{keff,keff2}. Among them, Liu Dong\cite{liu,liu2} applied PINNs to solve multiple neutron diffusion benchmark equations and demonstrated highly accurate predictive results. By utilizing FCN to capture neutron distribution, they achieved a neutron flux distribution solution with an accuracy as high as e-07 and successfully applied it to the search for $k_{eff}$. 

\begin{table*}[htbp]
  \centering
  \caption{Comparison Between PINN and Traditional Numerical Methods.}
  \label{tab:compare}
  \begin{tabular}{p{4cm}p{6cm}p{6cm}}
    \toprule
    \textbf{Feature} & \textbf{PINN} & \textbf{Traditional Numerical Methods} \\
    \midrule
    Calculation Method & Combines neural networks with physics equations & Uses analytical solutions or numerical methods \\
    Accuracy Dependency & Data noise level, physics model accuracy & Grid resolution, numerical methods \\
    Computational Efficiency & Low computational demand, training process needs hyperparameter optimization & High computational demand, requires significant computational resources \\
    Reliability & Robust to data noise and uncertainty & Sensitive to initial conditions, boundary conditions, and numerical parameters \\
    Model Complexity & Can handle complex nonlinear problems & Typically suitable for specific types and relatively simple problems \\
    Training Speed & Relatively fast training speed, But adjusting model hyperparameters is time-consuming & High computational complexity, longer runtime \\
    Grid Dependency & Grid-independent & Results are influenced by grid accuracy and partitioning \\
    Parameter Tuning & Requires tuning of network architecture and hyperparameters & Requires tuning of grid partitioning and solver parameters \\
    Parallel Computing & Efficient parallel computing & Has challenges in parallel computing \\
    Applicability & Suitable for complex nonlinear problems and data scarcity & Suitable for known fluid dynamics equations and stable boundary conditions \\
    \bottomrule
  \end{tabular}
\end{table*}

Moreover, FCNs are prone to gradient vanishing and can encounter non-convergence during training. When the gradient vanishes, NN parameters will not be updated and the network will find it hard to learn new knowledge\cite{vanish}. This may lead to the network having difficulty converging to an optimal solution, missing a large amount of information, and affecting the expressive power of the model. In the aspect of solving the neutron diffusion equation, the error in the predicted result will further make the critical judgment inaccurate. Furthermore, observation found that regions with large gradients will exhibit insufficient training under uniform sampling, especially when the number of sampling points is limited, leading to significant errors in regions with large gradients, thereby limiting further improvement in network accuracy. During solving the neutron diffusion equation, the limited accuracy may lead to inefficient parameter searches that greatly increase search time. 

Toward these challenges, recent studies propose adaptive sampling based on gradient information\cite{lu2021deepxde} and assigning adaptive weights to sample points in loss calculation\cite{mcclenny2023self,lossweight}. These enhancements improve the prediction performance of PINNs in regions with pronounced gradients and limited sampling points. Furthermore, the issue of gradient disappearance in FCNs remains unresolved. Researchers have explored replacing FCNs with other neural network techniques\cite{lawal2022physics}, such as CNNs\cite{fang2021high} and RNNs\cite{YangX}, to achieve better performance and more precise results. 

To this end, this paper proposes a novel framework called R$^2$-PINN, which combines the S-CNN architecture with the RAR method to solve neutron diffusion equations. The proposed model, by adding the gradient backpropagation path, effectively alleviates the gradient vanishing problem. Moreover, with a resample mechanism, the network can balance the loss between regions and reach a higher accuracy. R$^2$-PINN has been evaluated against the FCN for solving the neutron diffusion equation benchmark problems, which will be introduced in Section \ref{sec:2}, and shows that our method effectively suppresses loss function oscillation and achieves high-precision field prediction. In addition, our method significantly reduces the time required for an eigenvalue search, allowing us to obtain an accuracy of e-05 within just 10 minutes and obtain an accuracy of e-04 in only the 250s.

The rest of this paper is organized as follows. Section \ref{sec:2} introduces benchmark problems using in Section \ref{sec:4}, and Section \ref{sec:3} provides an introduction to the basic architectures, including the structure of S-CNN and RAR resample method and the overall R$^2$-PINN architecture. Section \ref{sec:4} provides multiple experiments to optimize hyperparameters and verify the superiority of the proposed model. Especially different search algorithms are compared for parameter search to try to reduce the search time. It also includes generalizability validation experiments using the same model for multiple benchmark problems. Then, Section \ref{sec:RD} analyzes and discusses the results of the experiment. Finally, Section \ref{sec:5} concludes our study and discusses future research directions.

\section{Problem Setup}\label{sec:2}
\subsection{One-dimensional reactor diffusion equation for a single energy group}\label{sec:21}
In this section, a single-group k-eigenvalue problem is introduced for criticality calculations. 
The largest value of $k$, known as the effective neutron multiplication factor or $k_{eff}$, needs to find out.
The equation of the single-group neutron diffusion model is commonly denoted as Eq.~\ref{eq:1}.

\begin{equation}\label{eq:1}
\begin{split}
\frac{1}{v} \frac{\partial \phi(r,E,t)}{\partial t} =&\nabla\cdot D\nabla\phi(r,E,t)\!-\!\Sigma_t(r,E)\phi(r,E,t)+\\
&x(E)\int_{0}^{\infty}\vartheta (E')\Sigma _f(r,E')\phi(r,E',t)dE'+\\
&S(r,E,t)\!+\!\int_{0}^{\infty }\Sigma_s(r,E'\!\to \!E)\phi(r,E',t)dE'  
\end{split}
\end{equation}
Where $\phi(r, E, t)$ denotes the neutron flux of energy group $E$ in the r-coordinate at the moment $t$, $v$ represents the neutron velocity, $D$ denotes the diffusion coefficient, $\vartheta$ represents neutrons per fission numbers, $x(E)$ represents prompt neutron spectra, $\Sigma_t$ represents total macroscopic cross section, $\Sigma_s$ represents macroscopic scattering cross section from group $E'$ to group $E$, $\Sigma_f$ represents fission macroscopic cross-section, and $S(r, E,t)$ represents the neutron source.

\begin{figure}[!htb]
\includegraphics
  [width=0.5\hsize]
  {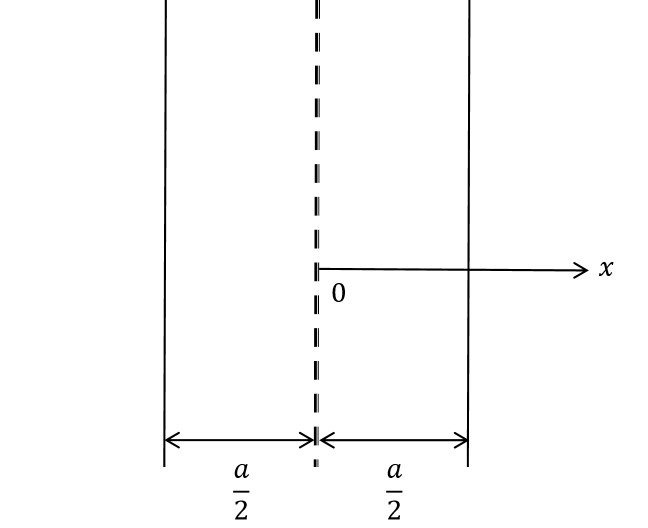}
\caption{Infinite Plate Reactor.}
\label{fig:1}
\end{figure}

In the absence of $S(r, E, t)$, an initial neutron flux density is symmetric along the x-axis. Eq.~\ref{eq:1} can be simplified into Eq.~\ref{eq:2} \cite{he}.

\begin{equation}\label{eq:2}
\frac{1}{Dv} \frac{\partial\phi(r,t) }{\partial t} =\nabla^2\phi(r,t)+\frac{k_\infty-1}{L^2}\phi(r,t) 
\end{equation}
Here, $k_\infty$ represents the infinite multiplication factor, and $L^2$ denotes the diffusion length.
Consider a uniform bare reactor\cite{NRP}, which is shaped as an infinite plate bare reactor with dimensions of infinite length and width and a thickness (including extrapolation distance) of $a$, as illustrated in Fig.~\ref{fig:1}. The analytical solution for the neutron flux can be obtained using the method of separation of variables, as given by Eq.~\ref{eq:3}.

\begin{equation}\label{eq:3}
\begin{split}
\phi(x,t)=&\sum_{n=1}^{\infty }\left [ \frac{2}{a} \int_{-\frac{2}{a} }^{\frac{2}{a}}\phi_0(x')cos\frac{(2n-1)\pi}{a}x'dx'\right ]\cdot\\
&cos\frac{(2n-1)\pi}{a}xe^{(k_n-1)t/l_n}  
\end{split}
\end{equation}

The critical condition for the bare reactor in the single-group approximation is given by Eq.~\ref{eq:4}.
\begin{equation}\label{eq:4}
k_{eff}=\frac{k_{\infty }}{1+L^2B^2} =1
\end{equation}
Here, $k_{eff}$ represents the effective neutron multiplication factor. When the system is in the criticality state, the neutron flux density distribution will satisfy the wave equation according to the fundamental eigenfunction corresponding to the minimum eigenvalue $B_g^2$, as given by Eq.~\ref{eq:5}.

\begin{equation}\label{eq:5}
\nabla^2\phi(r)+B_g^2\phi(r)=0
\end{equation}
If the composition of the system's materials (i.e., $k_\infty$ and $L^2$) is given, a unique critical size, denoted as $a_0$, makes $k_{eff}=1$. The critical size $a_0$ corresponds to the reactor being in a critical state. For reactor sizes larger than $a_0$, $k_{eff}>1$ indicates the reactor is in a supercritical state. Conversely, for reactor sizes smaller than $a_0$, $k_{eff}<1$, indicating the reactor is in a subcritical state\cite{NRA}. On the other hand, if the reactor size is given, it is always possible to find a fuel enrichment (material composition) that satisfies Eq.~\ref{eq:4} and ensures the reactor reaches criticality. When the system is in a critical state, the neutron flux density distribution within the reactor can be described as follows:

\begin{equation}\label{eq:stable}
\phi(x)=Acos\frac{\pi}{a} x
\end{equation}

\subsection{Two-dimensional reactor diffusion equation for a single energy group}\label{sec:22}
Based on \ref{sec:21}, consider the two-dimensional neutron diffusion equation,formulated as follows:

\begin{equation}\label{eq:2D_diffusion}
\frac{1}{Dv} \frac{\partial\phi(x,y,t) }{\partial t} =\nabla^2\phi(x,y,t)+\frac{k_\infty-1}{L^2}\phi(x,y,t) 
\end{equation}

By discretizing Eq.~\ref{eq:2D_diffusion} and approximating the Laplace operator and the partial derivatives, the equation can be converted to the following form:
\begin{equation}\label{eq:2D_diffusion_lisan}
\begin{split}
\frac{1}{Dv} \frac{\phi_{i,j}^{n+1} - \phi_{i,j}^{n}}{\Delta t} = &\frac{\phi_{i+1,j}^{n} - 2\phi_{i,j}^{n} + \phi_{i-1,j}^{n}}{\Delta x^2} +\\ &\frac{\phi_{i,j+1}^{n} - 2\phi_{i,j}^{n} + \phi_{i,j-1}^{n}}{\Delta y^2} +\\ &\frac{k_\infty - 1}{L^2} \phi_{i,j}^{n} 
\end{split}
\end{equation}
Where $\phi_{i,j}^{n}$ denotes the value of neutron flux at the spatial grid point $(i,j)$ at time $n$. $ \Delta t $ is the time step, $ \Delta x $ and $ \Delta y $ are the spatial steps in the $x$ and $y$ directions, respectively. $ D $ is the diffusion coefficient and $ v $ is the neutron velocity
$ k_\infty $ denote the infinite multiplication factor, and $ L $ is the diffusion length.

According to Eq.~\ref{eq:2D_diffusion_lisan}, the spatial domain meshes use the finite difference method to solve for the entire domain flux magnitude, and the numerically solved data are used as a test set to verify the model accuracy.

\subsection{Two-dimensional rectangular geometry multigroup multi-material diffusion problem}\label{sec:23}

In a nuclear reactor, neutron transport can be described by the multigroup diffusion theory. In this case, the fast and hot group neutron fluxes satisfy the following diffusion equations:

\begin{equation}\label{eq:fast_group}
-\nabla \cdot (D_1 \nabla \phi_1) + \Sigma_{r1} \phi_1 = \nu \Sigma_{f1} \phi_1
\end{equation}
\begin{equation}\label{eq:thermal_group}
-\nabla \cdot (D_2 \nabla \phi_2) + \Sigma_{r2} \phi_2 = \Sigma_{s1\to 2} \phi_1
\end{equation}

Where $D_1$ and $D_2$ are the diffusion coefficients of the fast and hot group neutrons, respectively.$\phi_1$ and $\phi_2$ are the fast group and thermal group neutron flux densities, respectively. $\Sigma_{r1}$ and $\Sigma_{r2}$ are the macroscopic cross sections of fast group and hot group neutrons, respectively.$\nu$ is the neutron yield. $\Sigma_{f1}$ is the fast group nuclear fission cross-section, and $\Sigma_{s1\to2}$ is the fast group to thermal group fission source term. 

\begin{figure}[!htb]
\includegraphics
  [width=0.55\hsize]
  {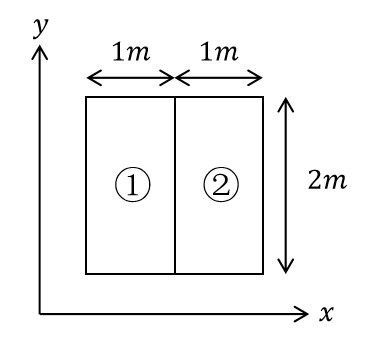}
\caption{Material Distribution\cite{liu2}.}
\label{fig:2}
\end{figure}

For pressurized water reactors,  the two-group diffusion are satisfied Eq.~\ref{eq:Problem31} and Eq.~\ref{eq:Problem32}, the example is divided into two different material regions, which is shown in Fig.~\ref{fig:2}, and the material parameters of each region are shown in Table.~\ref{tab:group_constants}.

\begin{equation}\label{eq:Problem31}
-D_{1} \nabla^{2} \phi_{1}(r)+\Sigma_{r1} \phi_{1}(r)\!=\!\frac{1}{k_{\mathrm{eff}}}\left[v \Sigma_{f,1}\phi_{1}(r)\!+\!v \Sigma_{f,2} \phi_{2}(r)\right]
\end{equation}

\begin{equation}\label{eq:Problem32}
-D_{2} \nabla^{2} \phi_{2}(r)+\Sigma_{a2} \phi_{2}(r)= \Sigma_{1 \to 2} \phi_{1}(r)
\end{equation}

\begin{table}[htbp]
\centering
\caption{Calculate Area Material Properties.}
\label{tab:group_constants}
\begin{tabular}{@{}lccccc@{}}
\toprule
 & \multicolumn{2}{c}{Material 1} & \multicolumn{2}{c}{Material 2} \\
\cmidrule(lr){2-3}\cmidrule(lr){4-5}
Energy Group & 1 & 2 & 1 & 2 \\
\midrule
$D_g(cm)$ & 1.268 & 0.1902 & 1.255 & 0.211 \\
$\Sigma_a$ ($cm^{-1}$) & 0.007181 & 0.07047 & 0.008252 & 0.1003 \\
$v\Sigma_f$ ($cm^{-1}$) & 0.004609 & 0.08675 & 0.004602 & 0.1091 \\
$\Sigma_{1\to 2}$ ($cm^{-1}$) & 0.02767 & - & 0.02533 & - \\
\bottomrule
\end{tabular}
\end{table}

Based on the multigroup diffusion theory, the distribution of neutron flux in the core is obtained by iteratively solving the discretized diffusion equation. The obtained dataset will be used as a test set in the experimental part to evaluate the model prediction accuracy.

\subsection{2D-IAEA benchmark problem}\label{sec:23}
The 2D-IAEA PWR benchmark problem is a two-dimensional statics problem with two neutron groups but with no delayed neutron precursors\cite{None}, which modeled by two-dimension two-group diffusion equations below:

\begin{equation}\label{eq:Two_group}
\left\{\begin{matrix}
 -D_{1}\nabla^2\phi_{1}\!+\!(\Sigma_{a,1}\!+\!\Sigma_{1\to 2}) \phi_{1}\!=\! \lambda \chi_1(\nu\Sigma_{f,1}\phi_{1}\!+\!\nu\Sigma_{f,2}\phi_{2})\\
-D_{2}\nabla^2\phi_{2}\!+\!\Sigma_{a,2}\phi_{2}\!-\!\Sigma_{1\to 2} \phi_{1}\!=\! \lambda \chi_2(\nu\Sigma_{f,1}\phi_{1}\!+\!\nu\Sigma_{f,2}\phi_{2})
\end{matrix}\right.
\end{equation}

\begin{figure}[!htb]
\includegraphics
  [width=0.75\hsize]
  {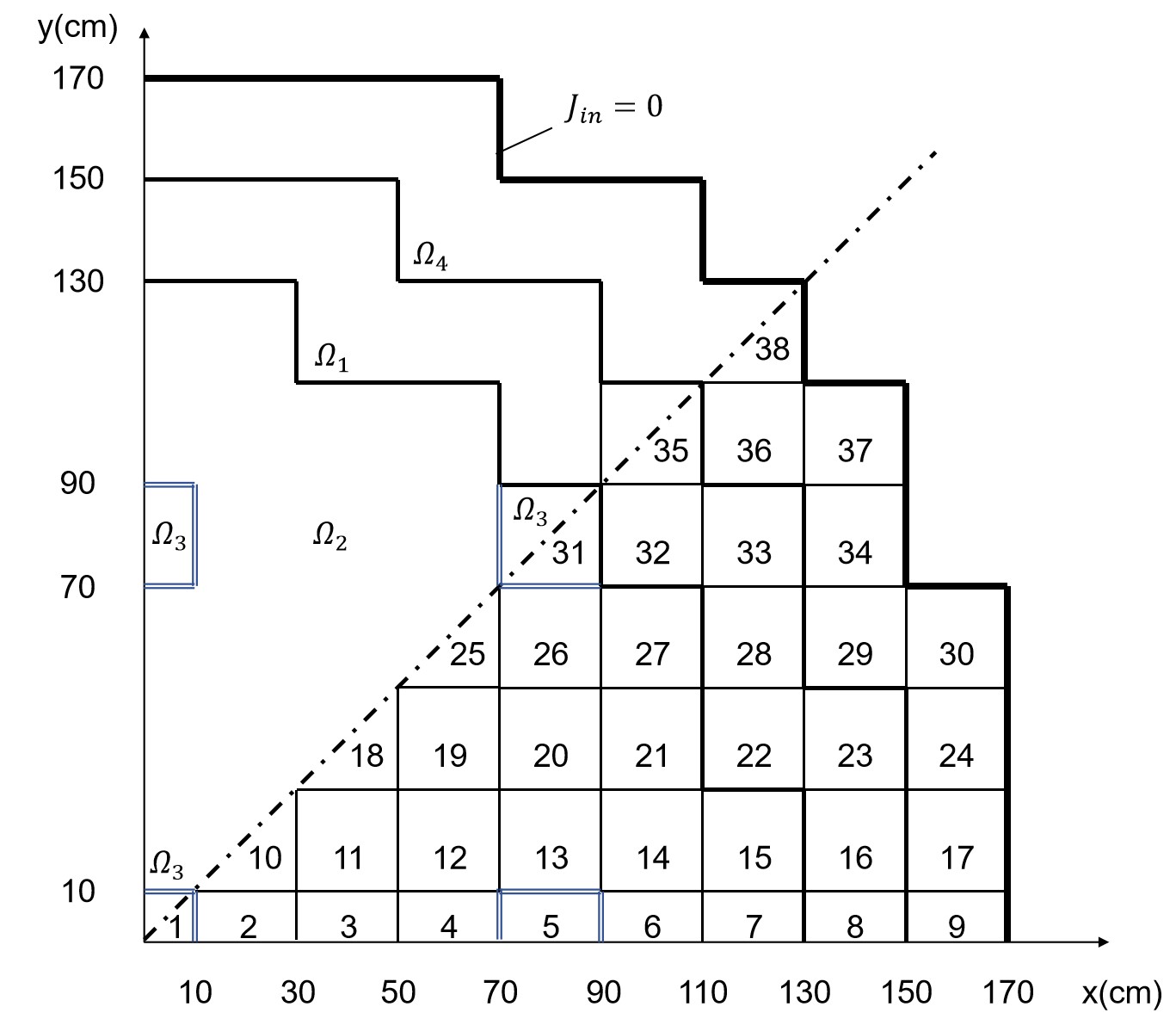}
\caption{Geometric layout of the 2D-IAEA benchmark problem\cite{None}.}
\label{fig:IAEA1}
\end{figure}

The reactor has a two-zone core containing 177 fuel assemblies which are 20 cm in width. The core is reflected radially by 20 cm of water. Due to the symmetry along the $x$ and $y$ axis,  This one quarter reactor domain is denoted by $\Omega$ and it is composed of four sub-regions of different
physical properties which are $\Omega_{1,2,3,4}$. The reactor is depicted in Fig.~\ref{fig:IAEA1}. Neumann boundary conditions are enforced on the left and the bottom boundaries. The group constants for this problem are given in Table.~\ref{tab:IAEA}.

\begin{table}[!htb]
\caption{Group constants of the 2D-IAEA benchmark.}
\label{tab:IAEA}
\begin{tabular*}{\columnwidth} {@{\extracolsep{\fill} } lccccc}
\toprule
 Region & Group & $D_g$(cm) & $\Sigma_{a,g}$(cm$^{-1}$) & $\nu \Sigma_{f,g}$(cm$^{-1}$) & $\Sigma_{1 \to 2}$(cm$^{-1}$)\\
\midrule
$\Omega_1$ & 1 & 1.5 & 0.010 & 0.000 & 0.02\\
  & 2 & 0.4 & 0.080 & 0.135 & \\
$\Omega_2$ & 1 & 1.5 & 0.010 & 0.000 & 0.02\\
  & 2 & 0.4 & 0.085 & 0.135 & \\
$\Omega_3$ & 1 & 1.5 & 0.010 & 0.000 & 0.02\\
  & 2 & 0.4 & 0.130 & 0.135 & \\
$\Omega_4$ & 1 & 2.0 & 0.000 & 0.000 & 0.04\\
  & 2 & 0.3 & 0.010 & 0.000 & \\
\bottomrule
\end{tabular*}
\end{table}

\section{Methods}\label{sec:3}
\subsection{PINN Loss Formulation}\label{sec:31}
In PINN, considering the general form of a parameterized and nonlinear PDE:

\begin{equation}\label{eq:7}
F(u,x,y,t,:)=0, (x,y)\in \Omega, t\in[0,T]
\end{equation}
Where $u$ represents the latent solution, and $\Omega$ represents the solution domain. This formula can express PDEs in almost all fields of mathematical physics. 

$N_f$ data points are sampled to measure physical consistency. This type of loss will be collectively referred to as the PDE loss in the following form:

\begin{equation}\label{eq:8}
Loss_{PDE}=\frac{1}{N_f} \sum_{i=1}^{N_f} F(u,x,y,t,:);x\in \Omega
\end{equation}

Then, through the introduction of initial loss and boundary loss due to the known initial and boundary conditions. $N_i$ and $N_b$ data points are collected for calculating $Loss_{Initial}$ and $Loss_{Boundary}$ respectively. Meanwhile, a few data need to be used for training. 

\begin{equation}\label{eq:initial}
Loss_{Initial}\!=\!\frac{1}{N_i} \sum_{i=1}^{N_i} (\phi_{truth}\!-\!\phi_{predict});t=0,x\in \Omega
\end{equation}
\begin{equation}\label{eq:boundary}
Loss_{Boundary}\!=\!\frac{1}{N_b} \sum_{i=1}^{N_b}(\phi_{truth}\!-\!\phi_{predict});x\in \partial\Omega
\end{equation}
\begin{equation}\label{eq:label}
Loss_{Data}\!=\!\frac{1}{N_d} \sum_{i=1}^{N_d}(\phi_{truth}\!-\!\phi_{predict})
\end{equation}
Incorporating a small amount of labeled data provides information about the correct order of magnitude, enabling the PINN to better calibrate its predictions and avoid unrealistic or divergent solutions. The inclusion of such labeled data greatly contributes to the stability and accuracy of the training process for PINNs.
By combining the losses above, networks can be penalized and the training process effectively constrained, thereby ensuring that the obtained solutions adhere more closely to the fundamental laws of physics. Therefore, the final network loss formulation is as follows:

\begin{align}
\label{eq:totalloss}
\begin{split}
Loss_{Total}\!=\!&Loss_{PDE}+\\
&(Loss_{Initial}\!+\!Loss_{Boundary}\!+\!Loss_{Data})\cdot w
\end{split}
\end{align}

The sampling ratio for the PDE loss, boundary loss, initial loss and data loss is approximately 30:10:10:1. Multiple losses can be balanced by adjusting weights $w$, preventing the network from prioritizing a single loss, especially when significant differences in magnitude between losses exist.

\subsection{S-CNN Architecture}\label{sec:32}

In solving the neutron diffusion equation, the performance of NN can be significantly affected by the vanishing gradient problem as the network depth increases. This problem can easily lead to training failure and make the result unreliable. Moreover, it greatly limits the growth of the number of layers, reduces the expressive power of the network, and limits the prediction accuracy of the network. Inspired by the effective alleviation of the gradient vanishing problem in ResNet architecture\cite{ResNet} in the image domain, which introduces the concept of residual learning, enables networks to learn residual mappings (the difference between input and desired output), making it easier to train very deep neural networks. The skip connection mechanism is introduced into the PINN architecture. 

The input sampling features are coordinates, namely $x$, $y$, and $t$, which are independent. So the experiment applies a separate filter to each feature. A one-dimensional convolution is used as the basis for each network layer. Each hidden layer follows the following formula:

\begin{equation}\label{eq:nn}
z_l=f_l(W_l z_{l-1}+b_l)
\end{equation}
Where $z_l$ denotes the hidden layer $l$ between the input and output layers, $W_l$ and $b_l$ are the weight and bias, respectively; $f_l(\cdot)$ denotes the activation function (e.g., tanh function.). 

\begin{figure}[!htb]
\includegraphics
  [width=1.0\hsize]
  {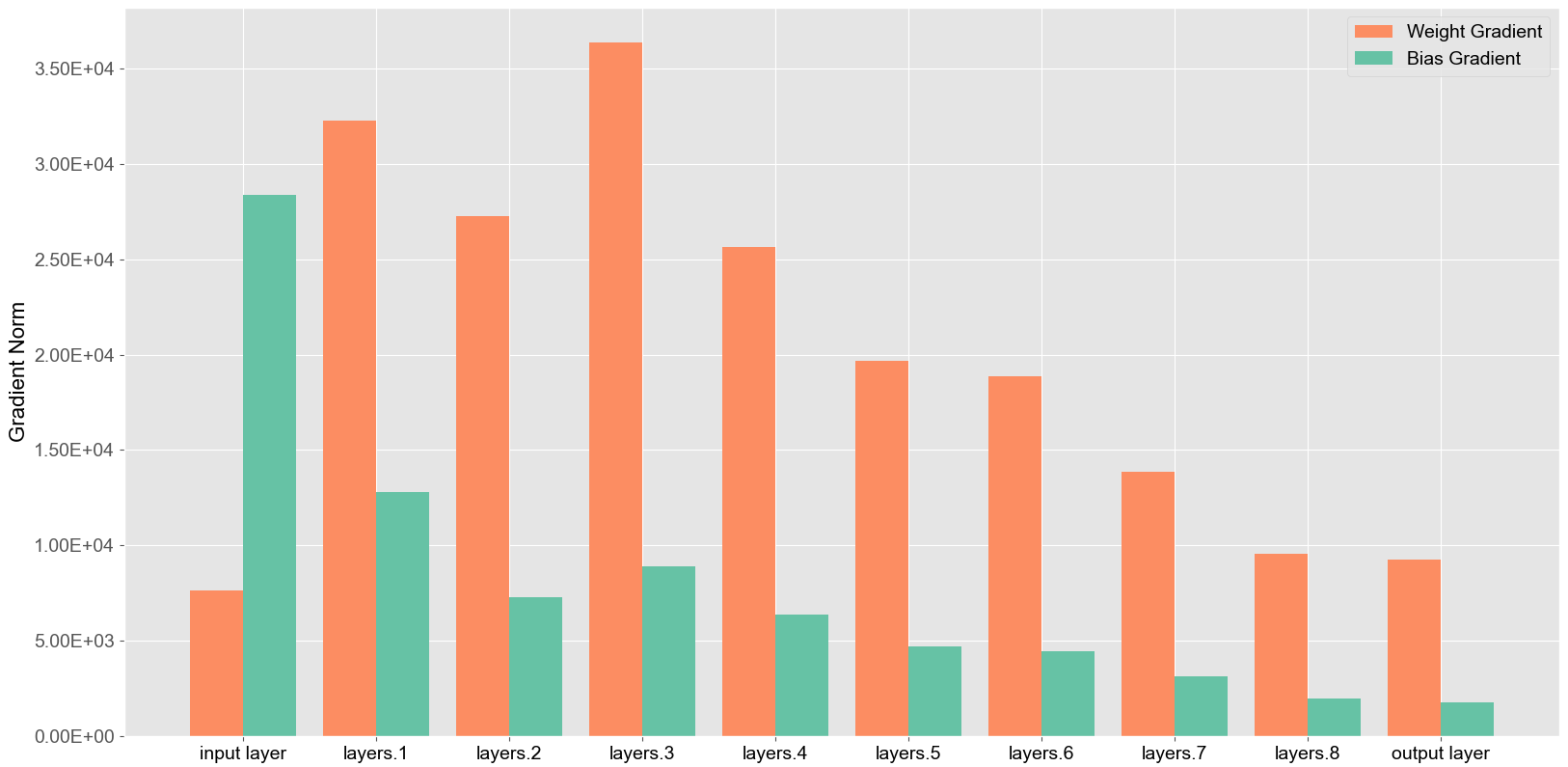}
\caption{Gradient Norm of Each Layer.}
\label{fig:GN}
\end{figure}

On this basis, some skip connections are added between different layers. The corresponding hidden layer can be represented as follows:

\begin{equation}\label{eq:Rnn}
z_l=f_l((W_l z_{l-1}+b_l)+z_{l-n-1}),l>n+1
\end{equation}
Where $n$ represents the skip distance, i.e., how many hidden layers are crossed. To determine where the skip connection should be added, the contribution of each layer to the training of the whole network is measured by calculating the gradient norm for each layer. Where the gradient norm is computed as follows\cite{MLiu}:

\begin{equation}\label{eq:gradient}
||g|| = sqrt(sum(g_i^2))
\end{equation}
For example, in the 10-layer network, the gradient contribution is calculated for each layer of the network, as shown in Fig.~\ref{fig:GN}.

When the gradient is more significant, it indicates that the layer's parameters can be easily updated. A too-large gradient will result in an unstable model, and a too-small gradient indicates that the layer may need to go through more iterations to reach the optimal state and may face the gradient vanishing problem. Thus, the network increases the shallow gradient paths by establishing shallow-to-deep skip connections, providing more gradient paths in shallow layers and avoiding vanishing gradients in deep layers. The detailed 10-layer S-CNN architecture used in Section \ref{sec:4} is shown in Table \ref{tab:S-CNN}, where B, C, and L represent the batch size, channel, and length.

\begin{table}[!htb]
\caption{S-CNN Structure.}
\label{tab:S-CNN}
\begin{tabular*}{\columnwidth} {@{\extracolsep{\fill} } lccc}
\toprule
 Layer Name & Input Size(B,C,L) & Output Size(B,C,L) & Param \\
\midrule
input layer & (None,2,1) & (None,26,1) & 78\\
conv1 layer & (None,26,1) & (None,26,1) & 702\\
conv2 layer & (None,26,1) & (None,26,1) & 702\\
conv3 layer & (None,26,1) & (None,26,1) & 702\\
conv4 layer & (None,26,1) & (None,26,1) & 702\\
conv5 layer & (None,26,1) & (None,26,1) & 702\\
conv6 layer & (None,26,1) & (None,26,1) & 702\\
conv7 layer & (None,26,1) & (None,26,1) & 702\\
conv8 layer & (None,26,1) & (None,26,1) & 702\\
output layer & (None,26,1) & (None,1,1) & 27\\
\bottomrule
\end{tabular*}
\end{table}

In the experimental section, the skip distance $n$ is set to 2 for larger spans of gradient propagation, which is the best parameter for shallow networks. when $n < 2$, the jump distance is insufficient, which may limit the network's ability to learn complex physical fields and does little to avoid the gradient vanishing problem, and when $n > 2$, the number of jumping layers is too large, which may pose a challenge to gradient propagation and thus increase the difficulty of optimizing the network. Experiments are conducted on S-CNN models with varying depths, and high-precision results are consistently obtained, thus validating the effectiveness of the residual module in addressing the gradient problem.

\subsection{RAR Mechanism}\label{sec:33}

Due to specific regions with large gradients in the overall solution domain, if sampled uniformly, the network suffers from inadequate fitting of local regions. This issue can be addressed by increasing the weights of specific sampling points, as shown in Eq.~\ref{eq:RARR1}, or by resampling using Eq.~\ref{eq:RARR2}.

\begin{equation}\label{eq:RARR1}
X_{new}=X_{raw}+w\cdot X_f
\end{equation}
\begin{equation}\label{eq:RARR2}
X_{new}=X_{raw}+X_{resample}
\end{equation}
Here, $X_{raw}$ represents the original dataset,$X_{new}$ represents the new dataset, $X_f$ represents the set of sampling points with more significant PDE residuals,$X_{resample}$ represents the set of resampling points, and $w$ represents the penalty weight.

Meanwhile, referring to the grid division of traditional numerical computation methods, finer-grained samples should be collected in regions with large gradients to improve the network's prediction accuracy in this region.

Based on this, consider using Eq.~\ref{eq:RARR2} to update our dataset, i.e., introducing RAR\cite{wu2023comprehensive} to improve the distribution of residual points during the training process of PINNs to address the bottleneck phenomenon that arises from the difficulty of reducing PDE residuals in certain regions, ultimately enhancing the predictive accuracy of the model.

By selectively sampling more points in regions where the PDE residuals are more significant, this approach allows the network to focus on challenging areas and adjust the sampling density accordingly, leading to improved learning and prediction capabilities. The method is particularly effective in capturing the complex behavior of the PDE solution and identifying sharp gradient regions.

The specific pseudo code for the RAR method is as follows:

\begin{algorithm}[H]
\SetAlgoLined
\caption{RAR Algorithm.}
\label{alg:rar_algorithm}
\SetKwInOut{Input}{Input}
\SetKwInOut{Output}{Output}
\Input{Set $S$ with randomly sampled initial points}
\Output{Updated set $S$}
Divide solution domain into $\alpha^2$ subintervals\;
Train PINN for $n$ iterations\;
\Repeat{the maximum number of iterations is reached or the total number of points reaches the limit}{
    Compute $Loss_{PDE}$ for the points in set $S$\;
    Calculate the average residual of each subinterval\;
    Randomly sample $S'$ from the subinterval with the highest average residual\;
    Update set $S$: $S=S\cup S'$\;
    Train PINN for $n$ iterations\;
}
\end{algorithm}

This algorithm divides the solution domain into $\alpha^2$ subintervals, with $\alpha$ subdivisions in both the $x$ and $t$ directions. The Latin Hypercube Sampling (LHS) method\cite{LHS} is used for random sampling, which ensures sample points cover the entire space without clustering or bias.

\begin{figure}[H]
	\centering  
	\subfigbottomskip=1pt 
	\subfigcapskip=-5pt 
	\subfigure[Before Resampling]{
		\includegraphics[width=0.47\linewidth]{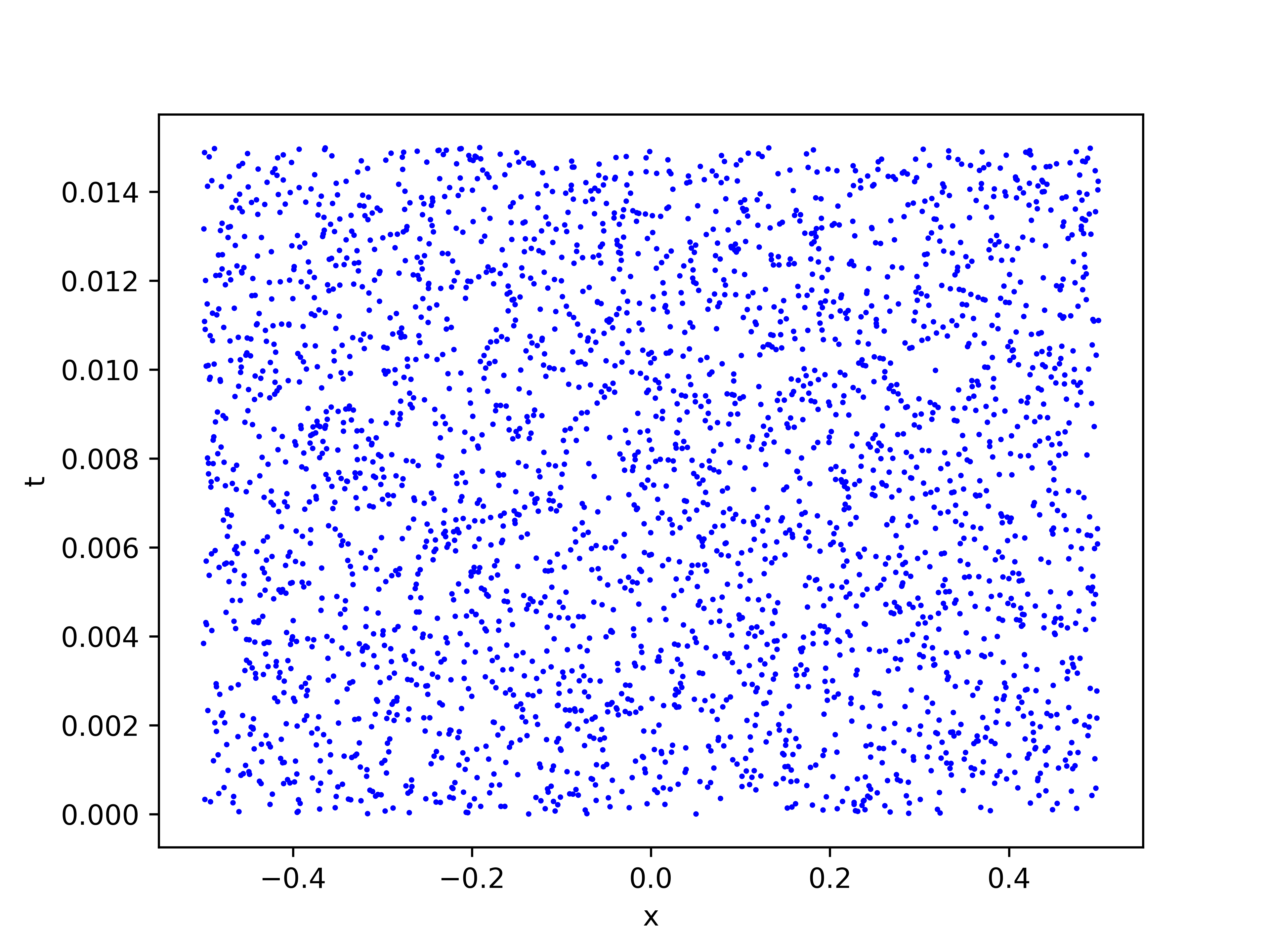} }
	\subfigure[After Resampling]{
		\includegraphics[width=0.47\linewidth]{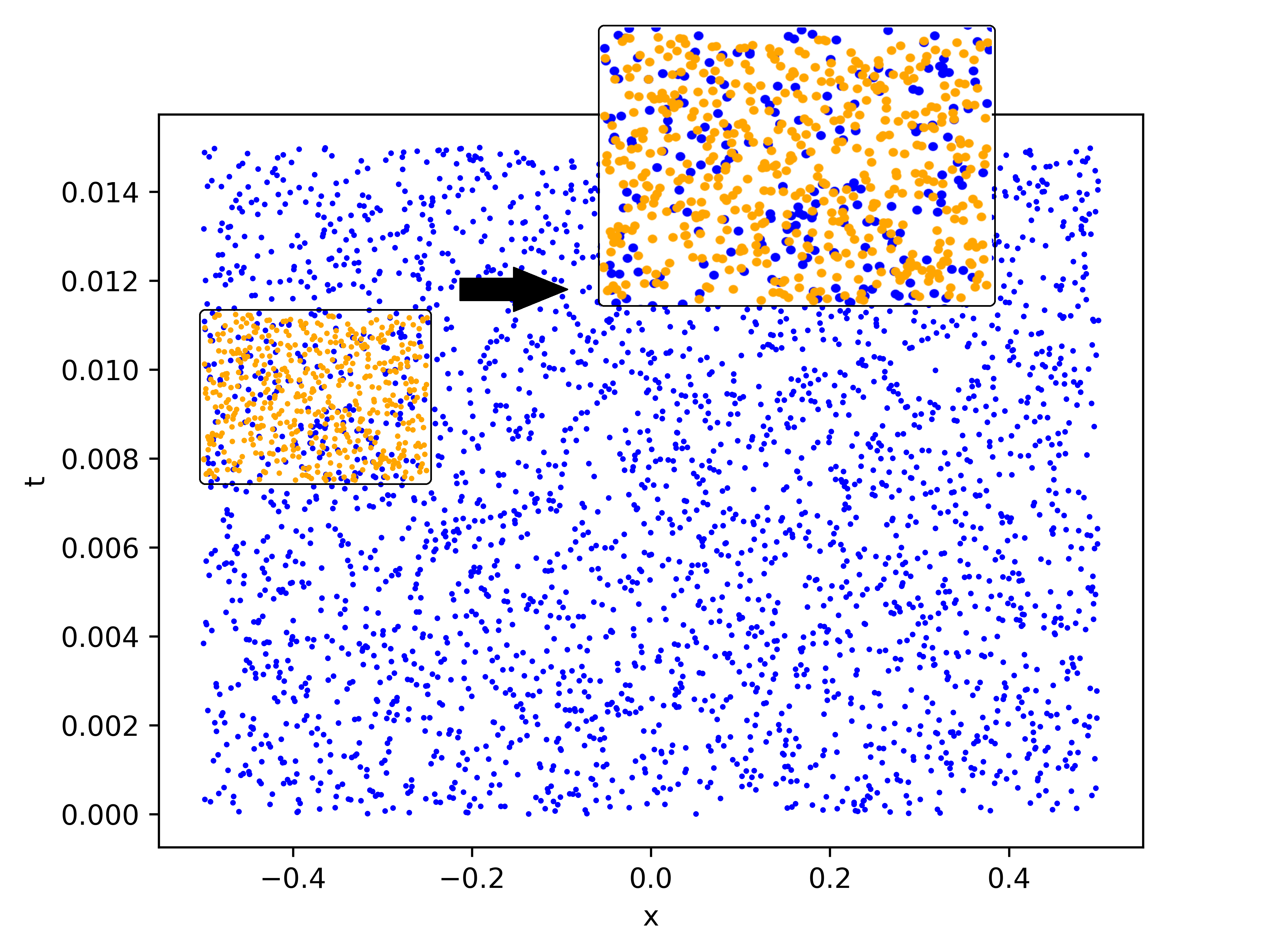} }
	\caption{PDE Points Distribution.}
 \label{fig:resample}
\end{figure}

As shown in Fig.~\ref{fig:resample}, by adaptively adding points in regions with more significant residuals, more intensive sampling has been conducted in one of the subdomains of the entire domain. This enables the network to capture the PDE solution's behavior better. This adaptive sampling approach improves the distribution of residual points and mitigates the bottleneck phenomenon, ultimately enhancing the accuracy and performance of the model and speeding up the model convergence.

\subsection{Proposed Framework:R$^2$-PINN}
Our paper proposes the R$^2$-PINN architecture, which combines the PINN structure, S-CNN, and RAR mechanism to solve PDEs with improved accuracy and computational efficiency.

\begin{figure*}[!htb]
\includegraphics
  [width=0.9\hsize]
  {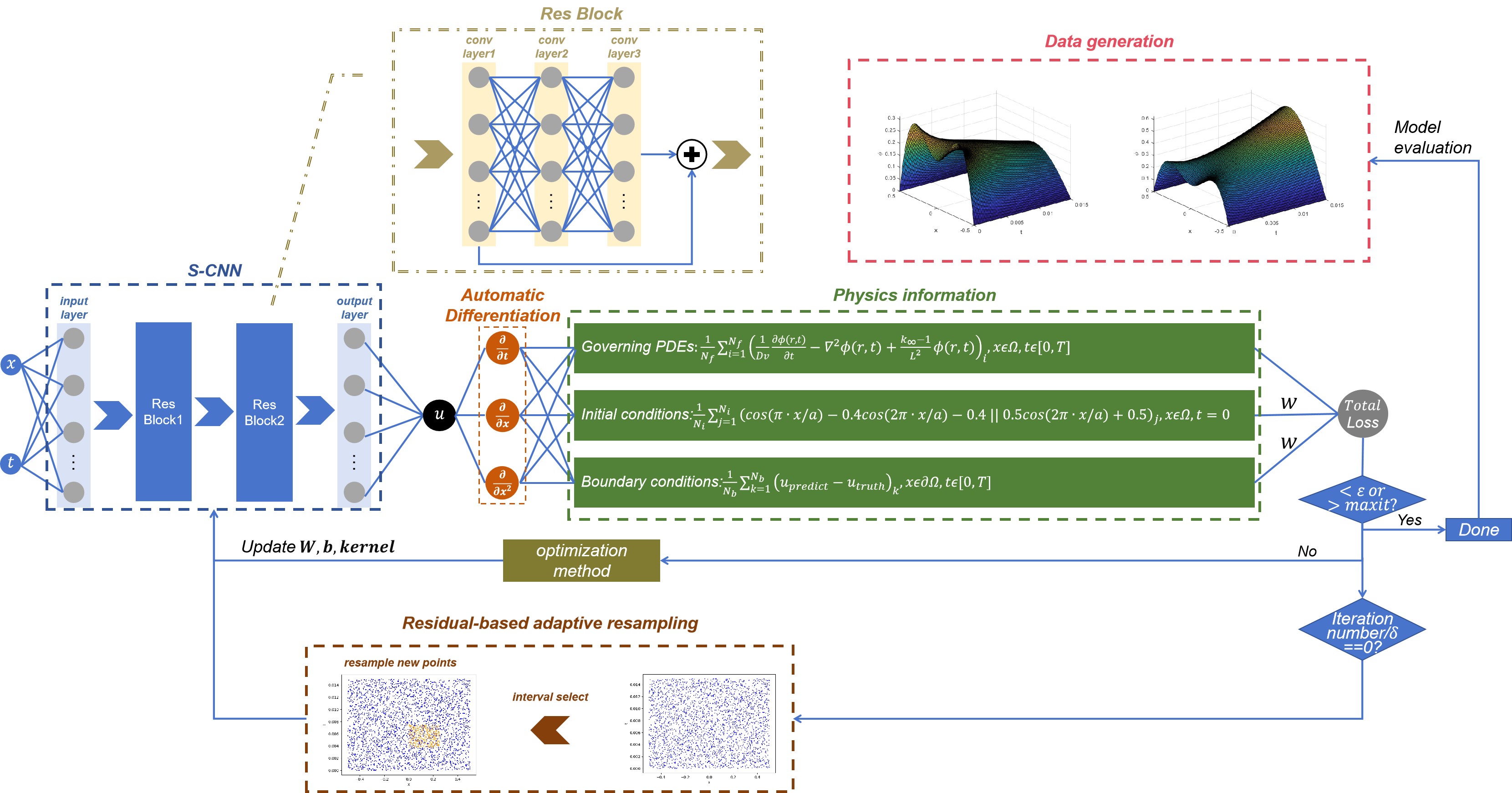}
\caption{Overall Architecture.}
\label{fig:overall}
\end{figure*}

The structure of R$^2$-PINN is shown in Fig.~\ref{fig:overall}. In R$^2$-PINN, the S-CNN is used as the backbone of the network, which can better capture the features of the PDE solution and improve the training efficiency. The RAR mechanism is employed to balance $Loss_{PDE}$ across different regions of the domain, ensuring that the network focuses on regions with high errors and adaptively refines the mesh. 

By combining the PINN structure, S-CNN, and RAR mechanism, R$^2$-PINN can effectively solve PDEs with improved accuracy and computational efficiency. It leverages the power of deep learning and adaptive refinement to capture complex features of the PDE solution, enabling a more precise representation of physical phenomena. Section \ref{sec:4} will verify the superiority of the proposed R$^2$-PINN network and each mechanism through a series of experiments.

\section{Experiments}\label{sec:4}
In Section \ref{sec:41}, the generation of datasets is introduced specifically for testing accuracy. Section \ref{sec:42}, through comparing S-CNN networks with different depths to validate the superiority of the improved PINN network architecture. Section \ref{sec:43} investigated how resampling affects the model's accuracy. In Section \ref{sec:44}, R$^2$-PINN is implemented to search for $k_{eff}$ and enhance search efficiency, discussed in Section \ref{sec:45}. Lastly, in Section \ref{sec:46}, the generalization capabilities of the improved PINN architecture model are tested. Furthermore, Sections \ref{sec:47} to \ref{sec:49} explore the generalizability of models in two-dimensional neutron diffusion, including multi-group and multi-material scenarios.

\subsection{Dataset Preparation}\label{sec:41}
\subsubsection{One-dimensional reactor diffusion equation for a single energy group}\label{sec:411}
According to Eq.~\ref{eq:3} and the assumption of symmetric initial conditions, two initial distributions are used for the dataset:
\begin{equation}\label{eq:12}
\begin{split}
&\phi_1(x,0)=cos(\pi \cdot x/a)-0.4cos(2\pi \cdot x/a)-0.4;\\
&\phi_2(x,0)=0.5cos(2\pi \cdot x/a)+0.5;\\
&x \in[-0.5,0.5]
\end{split}
\end{equation}
The specific numerical values used for the Eq.~\ref{eq:3} are as follows: $\nu =2.2 \times 10^3$m/s, $D=0.211 \times 10^{-2}$m, $L^2=2.1037 \times 10^{-4}\rm m^2$, $a=1$m. The boundary follows the Delicacy boundary and predicts the flux distribution for 0.015s.
By changing the $k_\infty$ parameter value, data can be obtained on how the flux varies with time for different parameter settings.

Our test dataset consists of 10000 data points with 100 grid points in both the $x$ and $t$ directions. The experimental part will test the model prediction accuracy on this dataset.

\subsubsection{Two-dimensional reactor diffusion equation for a single energy group}\label{sec:412}
To ensure that the boundary flux is 0 and the full domain flux is continuous, we set the initial flux distribution at the time t = 0 to the following equation:
\begin{equation}\label{eq:412}
\begin{split}
&\phi = (exp(-(x^2 + y^2)/20) - exp(-100))\\
&x \in[-10,10],y\in[-10,10].
\end{split}
\end{equation}

An iterative solution was performed using a time step $\Delta t$ of 0.001 with a grid number of 100 in the $x$,$y$,$t$ direction to solve for the flux distribution for 1 second, which was used as a dataset for testing the model.

\subsubsection{Two-dimensional rectangular geometry multigroup multi-material diffusion problem}\label{sec:413}

A source iteration method is used to solve for the dual-group neutron flux and is used to test the model's accuracy. The range of values of $x$,$y$ is shown in Fig.~\ref{fig:2}.And total dataset consists of 10000 data points with 100 grid points in both the $x$ and $y$ directions and the boundary follows the Delicacy boundary. 

\subsubsection{2D-IAEA problem}\label{sec:414}
The reference solution for the two-dimensional, two-group diffusion equations is obtained using the high-quality, general-purpose finite element solver FreeFem++\cite{FreeFem}. The eigenvalue problem is solved using tools from arpack++, the object-oriented version of the ARPACK eigenvalue package\cite{ARPACK}. The total number of samples in the dataset is 12286\cite{dataPINN}, using 76 of the data points as data feed into the network, and finally all the samples in the dataset are used to validate the model accuracy.

\subsection{S-CNN Depth and Kernel Size Ablation Experiment}\label{sec:42}
All parameters except the base network are kept consistent to compare the FCN architecture with the S-CNN architecture. The total number of sampled data points is 5000, with 3000 data points used to calculate the $Loss_{PDE}$, and 1000 data points sampled for the $Loss_{Initial}$ and $Loss_{Boundary}$, respectively. The network training does not involve feeding data to calculate $Loss_{data}$. Tanh activation function and LBFGS optimizer are used, and the Gaussian distribution random sampling method is used to initialize the network weights and biases. The number of hidden neurons per layer in the network is set to 26.

\begin{figure}
\centering
\subfigure[$\phi_0=\phi_1$]{
\includegraphics[width=2.5in]{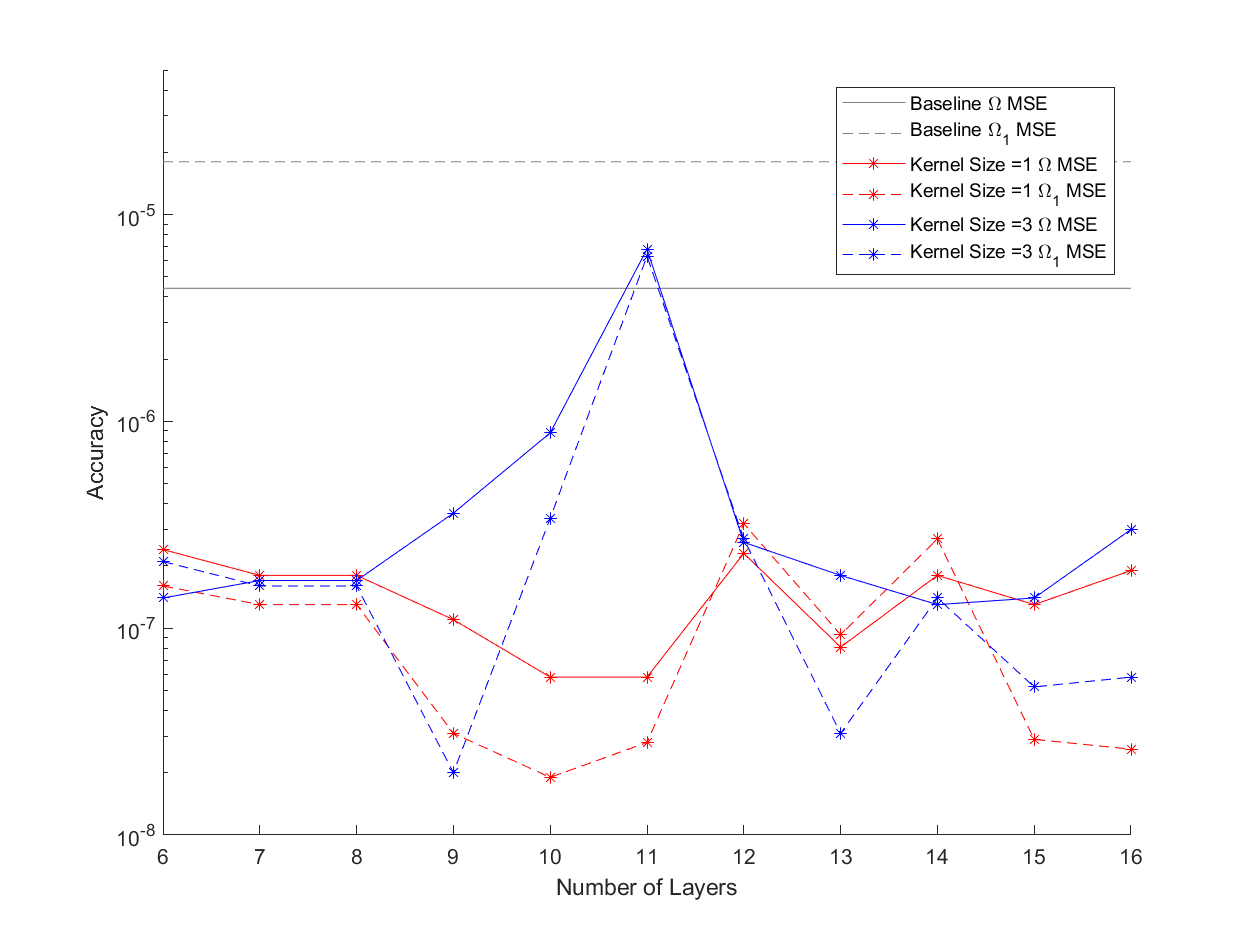}
}
\subfigure[$\phi_0=\phi_2$]{
\includegraphics[width=2.5in]{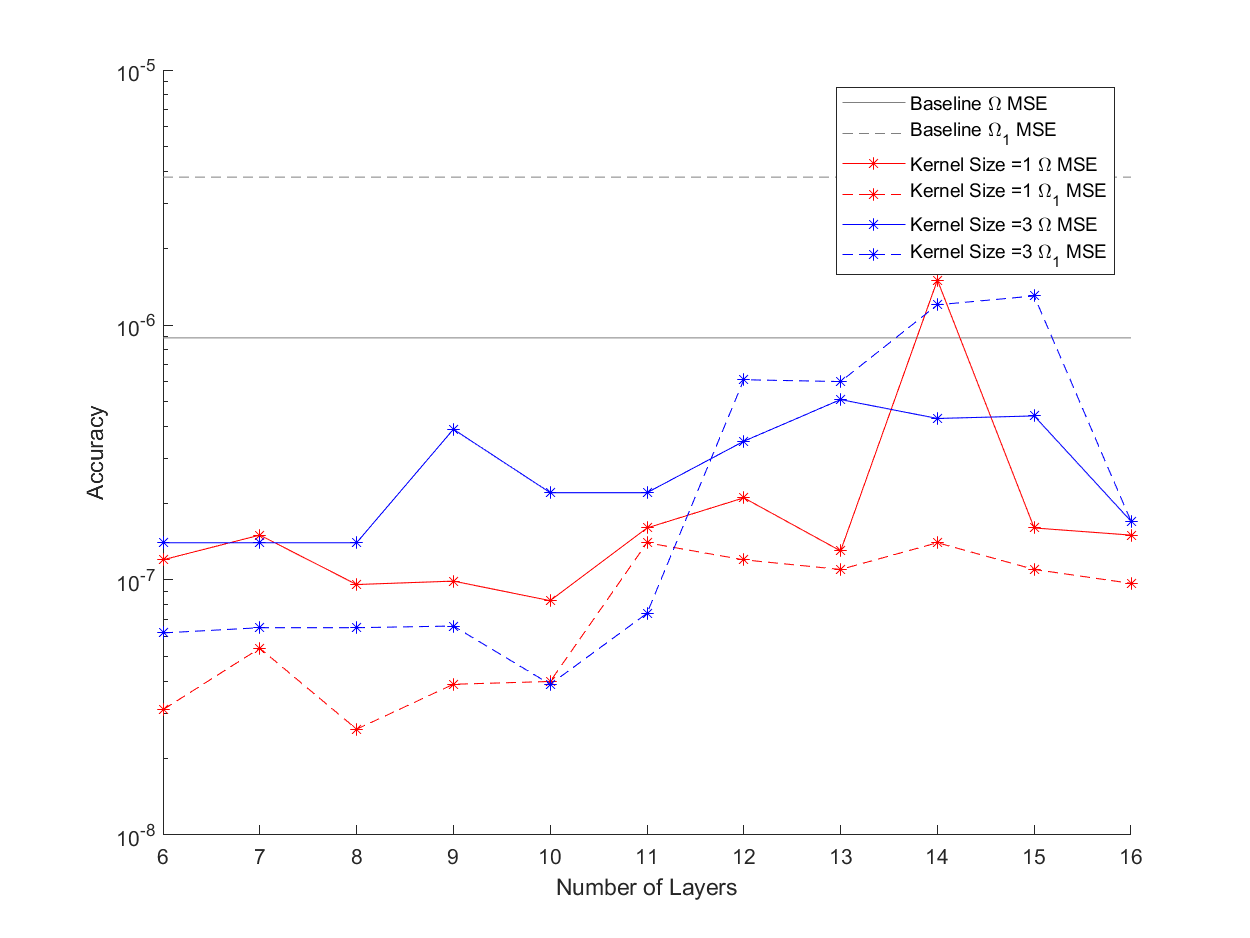}
}
\DeclareGraphicsExtensions.
\caption{Comparative Accuracy with Different Model Parameters.}
\label{fig:table3_vivid}
\end{figure}

To investigate the impact of layer numbers and kernel size on model performance, ablation experiments were conducted on S-CNN networks with different depths and different kernel sizes. For kernel size 1, the padding was set to 0; for kernel size 3, the padding was set to 1. The detailed experimental results are shown in Table \ref{tab:1} and Fig.~\ref{fig:table3_vivid}. 

\begin{table*}[!htb]
\caption{Ablation Study on S-CNN Depth and Kernel Size.}
\label{tab:1}
\begin{tabular*}{\linewidth}{@{\extracolsep{\fill}}lcccccccc}
\toprule
 & \multicolumn{4}{c}{k=1.0041 ($\phi_0=\phi_1$)} & \multicolumn{4}{c}{k=1.0001 ($\phi_0=\phi_2$)} \\
\cmidrule(lr){2-5} \cmidrule(lr){6-9}
& \multicolumn{2}{c}{Kernel Size=1} & \multicolumn{2}{c}{Kernel Size=3} & \multicolumn{2}{c}{Kernel Size=1} & \multicolumn{2}{c}{Kernel Size=3} \\
\cmidrule(lr){2-3} \cmidrule(lr){4-5} \cmidrule(lr){6-7} \cmidrule(lr){8-9}
S-CNN Layer & $\Omega$ MSE & $\Omega_1$ MSE & $\Omega$ MSE & $\Omega_1$ MSE & $\Omega$ MSE & $\Omega_1$ MSE & $\Omega$ MSE & $\Omega_1$ MSE \\
\midrule
Baseline & \num{8.9e-07} & \num{3.8e-06} & \num{8.9e-07} & \num{3.8e-06} & \num{4.4e-06} & \num{1.8e-05} & \num{4.4e-06} & \num{1.8e-05} \\ \addlinespace
6 & \num{1.2e-07} & \num{3.1e-08} & \num{1.4e-07} & \num{6.2e-08} & \num{2.4e-07} & \num{1.6e-07} & \num{1.4e-07} & \num{2.1e-07} \\ \addlinespace
7 & \num{1.5e-07} & \num{5.4e-08} & \num{1.4e-07} & \num{6.5e-08} & \num{1.8e-07} & \num{1.3e-07} & \num{1.7e-07} & \num{1.6e-07} \\ \addlinespace
8 & \num{9.6e-08} & \num{2.6e-08} & \num{1.4e-07} & \num{6.5e-08} & \num{1.8e-07} & \num{1.3e-07} & \num{1.7e-07} & \num{1.6e-07} \\ \addlinespace
9 & \num{9.9e-08} & \num{3.9e-08} & \num{3.9e-07} & \num{6.6e-08} & \num{1.1e-07} & \num{3.1e-08} & \num{3.6e-07} & \num{2.0e-08} \\ \addlinespace
10 & \num{8.3e-08} & \num{4.0e-08} & \num{2.2e-07} & \num{3.9e-08} & \num{5.8e-08} & \num{1.9e-08} & \num{8.8e-07} & \num{3.4e-07} \\ \addlinespace
11 & \num{1.6e-07} & \num{1.4e-07} & \num{2.2e-07} & \num{7.4e-08} & \num{5.8e-08} & \num{2.8e-08} & \num{6.8e-06} & \num{6.3e-06} \\ \addlinespace
12 & \num{2.1e-07} & \num{1.2e-07} & \num{3.5e-07} & \num{6.1e-07} & \num{2.3e-07} & \num{3.2e-07} & \num{2.6e-07} & \num{2.7e-07} \\ \addlinespace
13 & \num{1.3e-07} & \num{1.1e-07} & \num{5.1e-07} & \num{6.0e-07} & \num{8.1e-08} & \num{9.3e-08} & \num{1.8e-07} & \num{3.1e-08} \\ \addlinespace
14 & \num{1.5e-06} & \num{1.4e-07} & \num{4.3e-07} & \num{1.2e-06} & \num{1.8e-07} & \num{2.7e-07} & \num{1.3e-07} & \num{1.4e-07} \\ \addlinespace
15 & \num{1.6e-07} & \num{1.1e-07} & \num{4.4e-07} & \num{1.3e-06} & \num{1.3e-07} & \num{2.9e-08} & \num{1.4e-07} & \num{5.2e-08} \\ \addlinespace
16 & \num{1.5e-07} & \num{9.7e-08} & \num{1.7e-07} & \num{1.7e-07} & \num{1.9e-07} & \num{2.6e-08} & \num{3.0e-07} & \num{5.8e-08} \\ 
\bottomrule
\end{tabular*}
\end{table*}

In Table \ref{tab:1}  and Fig.~\ref{fig:table3_vivid}, $\Omega$ MSE refers to the MSE score over the entire domain, and $\Omega_1$ MSE refers to the MSE score at $t=0.015s$, which is calculated separately to assess the model's ability to extrapolate in time. Baseline refers to the result of FCN. 

Based on Table \ref{tab:1} and Fig.~\ref{fig:table3_vivid}, it can be observed that when layer numbers are lower than 10, the accuracy shows a positive correlation with layer numbers. But as the layer number further increases, the accuracy cannot reach higher. Hence, it is better to choose fewer layer numbers to make the NN train faster and achieve high accuracy. This is because the expressive ability of the neural network increases as the number of layers increases, but beyond a certain threshold, the increase in the number of layers does not significantly improve the expressive ability of the network; the increase in the number of layers only increases the training time consumed, and the model suffers from overfitting problems. Moreover, Because the input features are coordinates and have no relation to each other, the kernel size set to 1 can consider each channel as an independent feature to tackle, and this operation is more in line with reality. So, a kernel size set to 1 can achieve higher accuracy than a kernel size set to 3. Based on the above analysis, the 10-layer S-CNN with a kernel size of 1 exhibits the best predictive accuracy with a minimum number of parameters. Consequently, this configuration will be utilized for subsequent experiments. 

\subsection{Ablation Experiment on Resampling Parameters in S-CNN}\label{sec:43}
According to the RAR algorithm, the resampling granularity is set to $\alpha$ to define the granularity of the subdomains in the solution domain. The solution domain is divided into $\alpha$ subintervals along the $x$ and $t$ dimensions, resulting in $\alpha^2$ subintervals. Following the RAR algorithm, every 1000 epochs, the network calculates and compares the MSEs of each subdomain and performs resampling in the subinterval with the largest MSE. The number of resampling points is denoted as $m$.

The initial PDE sampling point is set to 3000 to avoid excessive sampling and training time. Moreover, 2000 data points are sampled, with 1000 points to calculate the initial loss and 1000 points to calculate the boundary loss. The maximum number of PDE samples is set to 5000, and resampling stops when the total number of PDE samples reaches 5000 after multiple iterations. The LHS method is used for each sampling.

Ablation experiments are conducted from two dimensions: the resampling granularity and the number of resampling points while maintaining the same S-CNN architecture (10 layers) and other hyperparameters. The S-CNN network is compared with an FCN network (baseline) using the same initial condition, i.e. $\phi_0=\phi_1$. The detailed experimental results are shown in Table \ref{tab:2} and Table \ref{tab:3}.

\begin{table}[htb]
\caption{Ablation Study on Resampling Numbers.}
\label{tab:2}
\begin{tabular*}{\linewidth}{@{\extracolsep{\fill}}lcc}
\toprule
Resample Numbers $m$ & {$\Omega$ MSE} & {$\Omega_1$ MSE} \\
\midrule
Baseline  & \num{8.9e-07} & \num{3.8e-06} \\
0  & \num{9.1e-08} & \num{2.1e-08} \\
100  & \num{1.3e-07} & \num{8.4e-08} \\
200  & \num{9.5e-08} & \num{9.8e-09} \\
300  & \num{1.3e-07} & \num{5.4e-08} \\
400  & \num{9.8e-08} & \num{4.1e-08} \\
500  & \num{8.0e-08} & \num{4.9e-09} \\
600  & \num{1.3e-07} & \num{3.1e-08} \\
700  & \num{1.0e-07} & \num{2.0e-08} \\
800  & \num{3.9e-08} & \num{1.8e-08} \\
\bottomrule
\end{tabular*}
\end{table}

\begin{table}[htb]
\caption{Ablation Study on Resampling Granularity.}
\label{tab:3}
\begin{tabular*}{\linewidth}{@{\extracolsep{\fill}}lcc}
\toprule
Resample Granularity $\alpha$ & {$\Omega$ MSE} & {$\Omega_1$ MSE} \\
\midrule
Baseline  & \num{8.9e-07} & \num{3.8e-06} \\
2  & \num{8.0e-08} & \num{4.9e-09} \\
3  & \num{1.5e-07} & \num{8.5e-08} \\
4  & \num{1.2e-08} & \num{9.1e-08} \\
5  & \num{6.4e-07} & \num{3.5e-08} \\
6  & \num{1.0e-07} & \num{8.8e-08} \\
\bottomrule
\end{tabular*}
\end{table}

From Table \ref{tab:2} and Table \ref{tab:3}, it is evident that under different sampling hyperparameters, our network consistently achieves a better $\Omega_1$ MSE result that is two orders of magnitude higher than the FCN baseline. When using the optimal hyperparameters, R$^2$-PINN can achieve an e-08 or even e-09 accuracy. This indicates that our method has a significant advantage in determining whether the flux values reach a steady state at a specific moment. This advantage will also be demonstrated during the $k_{eff}$ search in Section \ref{sec:44}.

\begin{figure}[!htb]
\includegraphics
  [width=0.8\hsize]
  {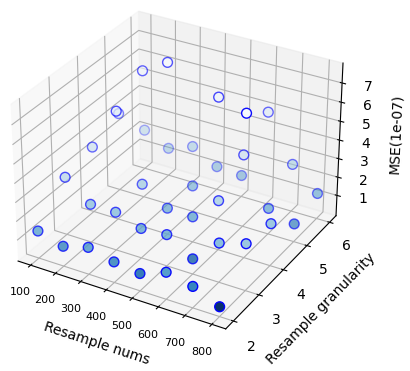}
\caption{Resample Parameter 3D Visualization.}
\label{fig:3d}
\end{figure}

\begin{figure}[H]
	\centering  
	\subfigbottomskip=1pt 
	\subfigcapskip=-5pt 
	\subfigure[$m$ Comparison]{
		\includegraphics[width=0.47\linewidth]{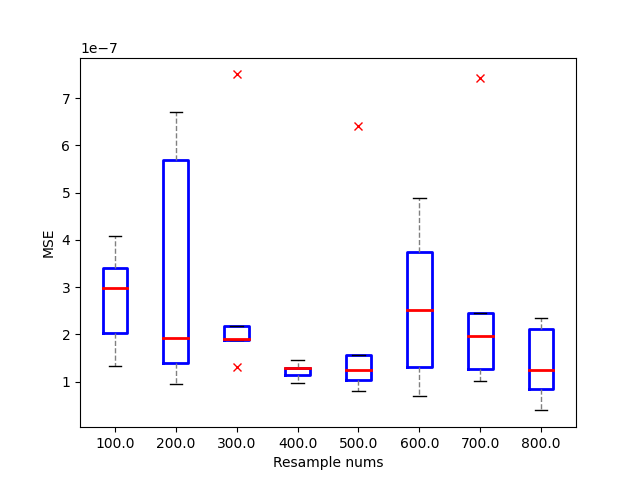} }
	\subfigure[$\alpha$ Comparison]{
		\includegraphics[width=0.47\linewidth]{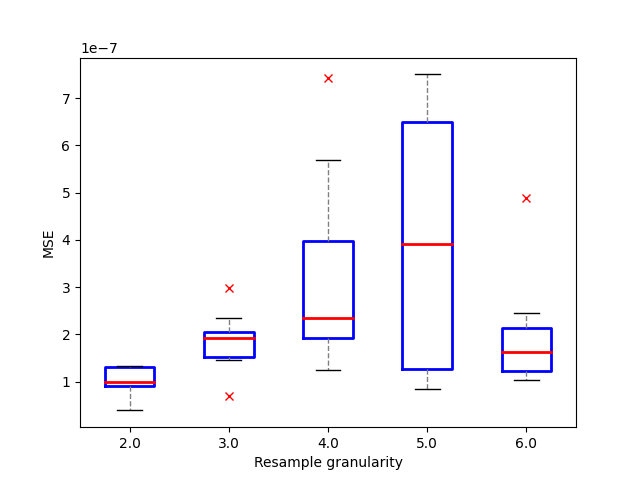} }
	\caption{MSE Comparison Between Parameters.}
 \label{fig:34}
\end{figure}

Furthermore, the results are obtained by adopting another initial condition where $\phi_0=\phi_2$. Then, a boxplot analysis on the resampling hyperparameters for all the results, which are shown in Fig.~\ref{fig:3d} and Fig.~\ref{fig:34}. It can be observed that when the number of resamples is 500 and the resample granularity is set to 2, the model can reach the best accuracy.

Given the introduction of the RAR mechanism, it is necessary to consider whether there is a large time overhead. Hence, time comparisons are performed for different numbers of samples, the results of which are shown in Fig. ~\ref{tab:Resample_time}.  It can be observed that when the number of resamplings is set to 500, the training time required by the model is significantly reduced as compared to setting the number of resamplings to zero. By calculating the time consumed per cycle, we find that the trend does not increase significantly. Thus, setting the appropriate number of resamplings effectively reduces the overall network training time. This shows that the RAR method has an increase in the number of sampling points, leading to a different density of samples in each region as well as a relatively large number of sampling points in complex regions, which greatly contributes to the convergence speed of the network.

\begin{table}[htb]
\caption{Training Performance with Different Resample Numbers.}
\label{tab:Resample_time}
\begin{tabular*}{\linewidth}{@{\extracolsep{\fill}}lccc}
\toprule
$m$ & Training Epochs	&Training Time (s) &	Avg. Time per Epoch(s)\\
\midrule
0     &3262 	&275.28 &0.084\\
100   &3875	&300.36	&0.078\\
200   &1005	&84.25	&0.084\\
300   &2004	&174.32	&0.087\\
400   &2544 &164.16	&0.065\\
\textbf{500}   &\textbf{1004} &\textbf{84.48}	&\textbf{0.084}\\
600   &3004 &240.74	&0.080\\
700   &3004 &287.95	&0.096\\
800   &2005 &140.02	&0.070\\
\bottomrule
\end{tabular*}
\end{table}

\begin{figure}[H]
	\centering  
	\subfigbottomskip=1pt 
	\subfigcapskip=-5pt 
	\subfigure[$k_\infty=1.0001$]{
		\includegraphics[width=0.47\linewidth]{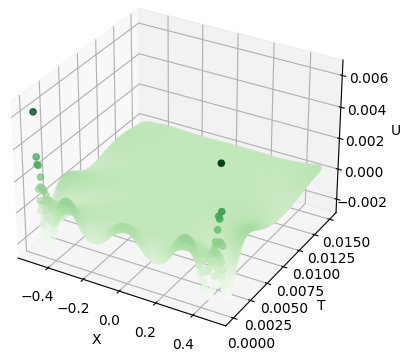} }
	\subfigure[$k_\infty=1.0041$]{
		\includegraphics[width=0.47\linewidth]{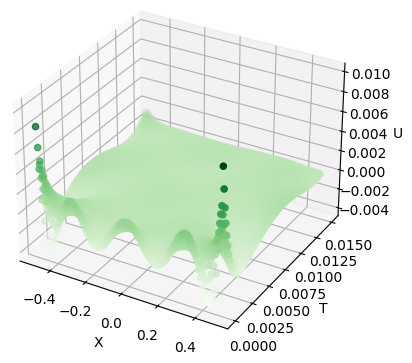} }
	\caption{Error Field.}
 \label{fig:EF}
\end{figure}

\begin{figure}[H]
	\centering  
	\subfigbottomskip=1pt 
	\subfigcapskip=-5pt 
	\subfigure[$k_\infty=1.0001$]{
		\includegraphics[width=0.47\linewidth]{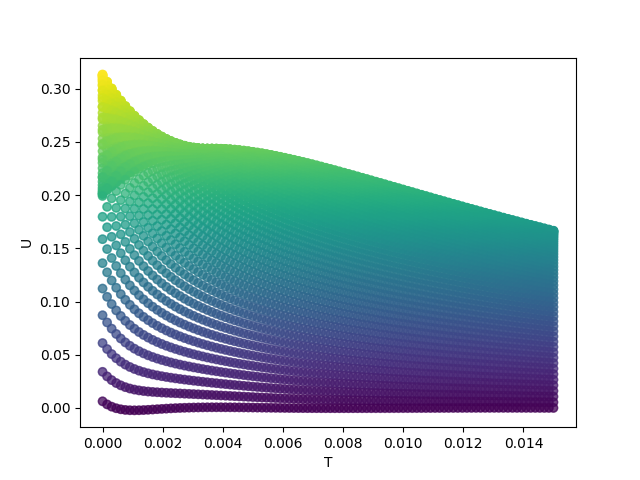} }
	\subfigure[$k_\infty=1.0041$]{
		\includegraphics[width=0.47\linewidth]{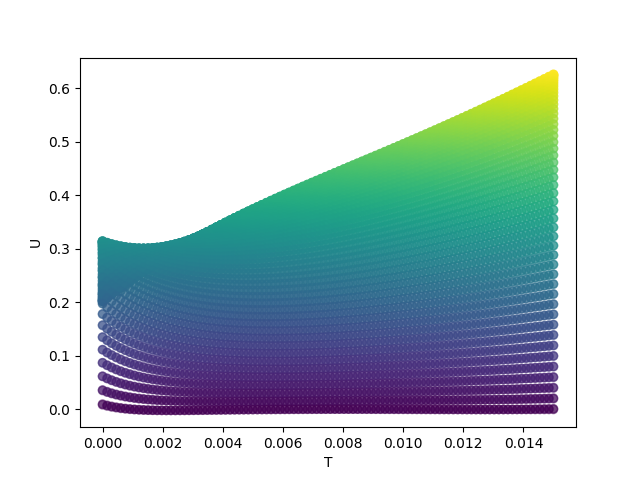} }
	\caption{Predict Result.}
 \label{fig:pr}
\end{figure}

The model uses the optimal resampling parameter configuration to predict the flux distribution under different $k_\infty$ values where  $\phi_0=\phi_1$, the error plot is shown in Fig.~\ref{fig:EF}. Except for the relatively large errors at $t=0$, the errors in other regions are relatively flat. Notably, as time increases, the errors show a slight increase. In addition, Fig.~\ref{fig:pr} shows the distribution of predicted values by the model. It can be observed that when t approaches 0, there is a specific deviation between the data in the boundary region and 0. This guides us to appropriately increase the weights of boundary loss and initial condition loss. 

\begin{figure}[H]
	\centering  
	\subfigbottomskip=1pt 
	\subfigcapskip=-5pt 
	\subfigure[Train Loss($k_\infty=1.0001$)]{
		\includegraphics[width=0.47\linewidth]{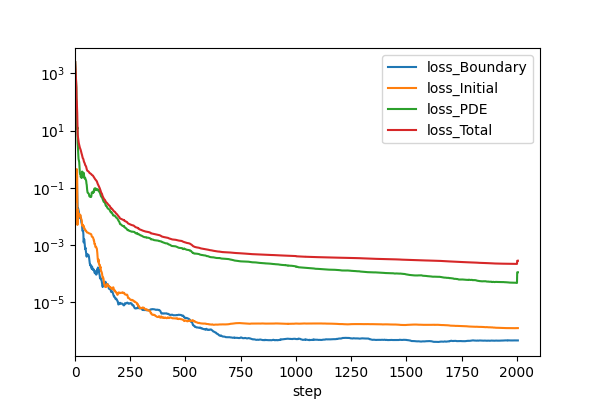} }
	\subfigure[Train Loss($k_\infty=1.0041$)]{
		\includegraphics[width=0.47\linewidth]{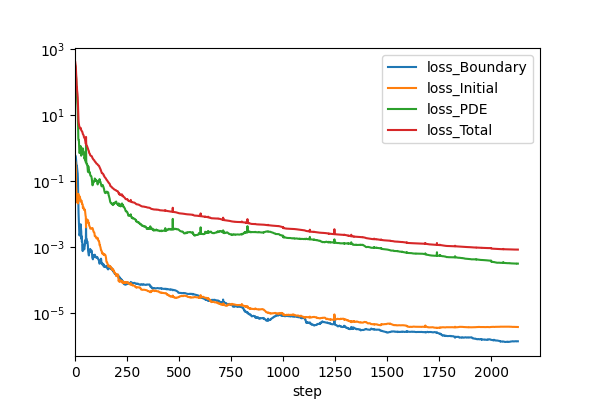} }
        \subfigure[Test Loss]{
		\includegraphics[width=0.47\linewidth]{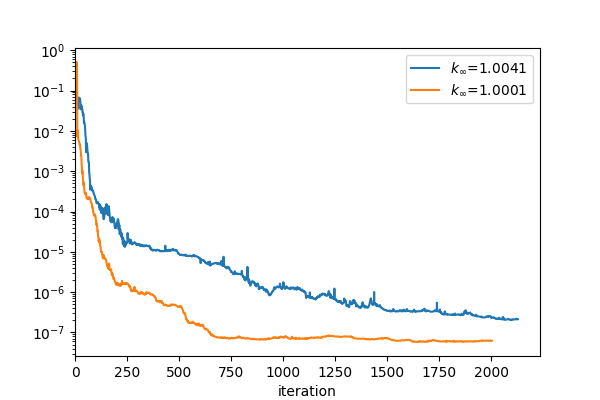} }
	\caption{Loss Over Iterations.}
 \label{fig:Train}
\end{figure}

To evaluate the model's performance and measure the difference between losses, Fig.~\ref{fig:Train} show the training and test losses separately. In Fig.~\ref{fig:Train}, $Loss_{PDE}$ is bigger than $Loss_{Boundary}$ and $Loss_{Initial}$, this may lead to $Loss_{PDE}$ dominate the optimization process and other losses cannot be optimized well. To address this issue, $w$ in Eq.~\ref{eq:totalloss} is set to 100 to impose greater penalties on the boundary and the initial region error. R$^2$-PINN can converge in just 2000 epochs.  

Based on experiment results in Section \ref{sec:42} and Section \ref{sec:43}, an optimized S-CNN architecture with optimized RAR mechanism is used to compose the R$^2$-PINN to further used to search $k_{eff}$, which will conduct in Section \ref{sec:44} and Section \ref{sec:45}.

\subsection{$k_{eff}$ Search with R$^2$-PINN}\label{sec:44}
For a given geometric shape and volume of the reactor core, $k_\infty$ and $L^2$ can be adjusted by modifying the reactor core size or changing the composition of materials within the reactor so that $k_{eff}$ equals 1. When the system reaches a steady state after a sufficient period, the neutron flux density follows the distribution described by Eq.~\ref{eq:stable}, and the reactor will be in a critical state.

\begin{figure}[H]
	\centering  
	\subfigbottomskip=2pt 
	\subfigcapskip=-5pt 
	\subfigure[$\phi_0=\phi_1$, $k_\infty=1.0001$]{
		\includegraphics[width=0.48\linewidth]{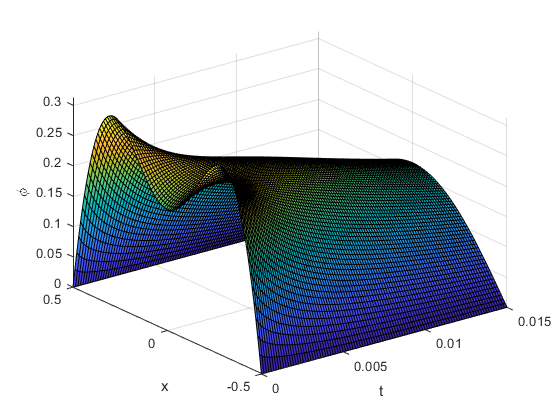}}
	\subfigure[$\phi_0=\phi_1$, $k_\infty=1.0041$]{
		\includegraphics[width=0.48\linewidth]{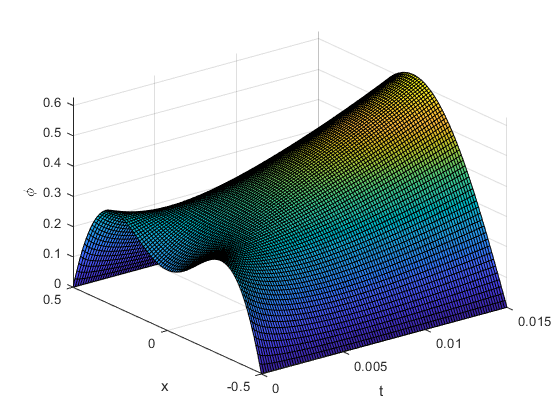}}
	  \\
	\subfigure[$\phi_0=\phi_2$, $k_\infty=1.0001$]{
		\includegraphics[width=0.48\linewidth]{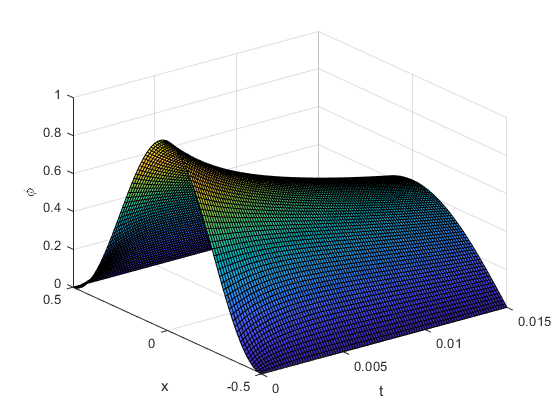}}
	\subfigure[$\phi_0=\phi_2$, $k_\infty=1.0041$]{
		\includegraphics[width=0.48\linewidth]{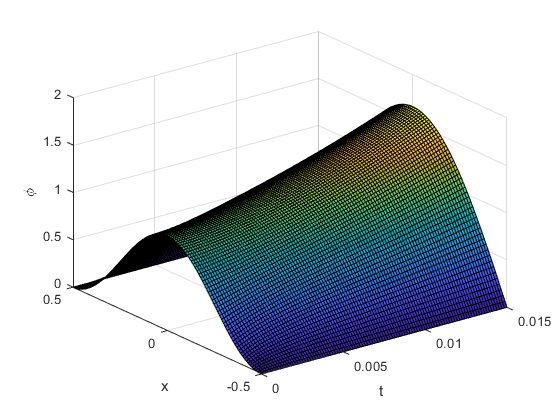}}
	\caption{$\phi$ Distribution.}
 \label{fig:4}
\end{figure}

In this experiment, maintain the $L^2$ parameter size unchanged and train different networks by adjusting $k_\infty$ to predict the evolution of neutron flux at this parameter. By continuously adjusting the value of the $k_\infty$ parameter, try to adjust $k_{eff}$ to 1, reaching a critical state. Initially, the parameter search range is set to [1.0001, 1.0041]. It has been verified that when $k_\infty$ is 1.0001, $k_{eff}<1$, indicating a subcritical state, where $\phi(x,t)$ exponentially decays with time t. When $k_\infty$ is 1.0041,$k_{eff}>1$ indicates a supercritical state, where $\phi(x,t)$ continuously increases. The specific distribution figure is shown in Fig.~\ref{fig:4}.

\begin{figure*}[!htb]
\includegraphics
  [width=0.54\hsize]
  {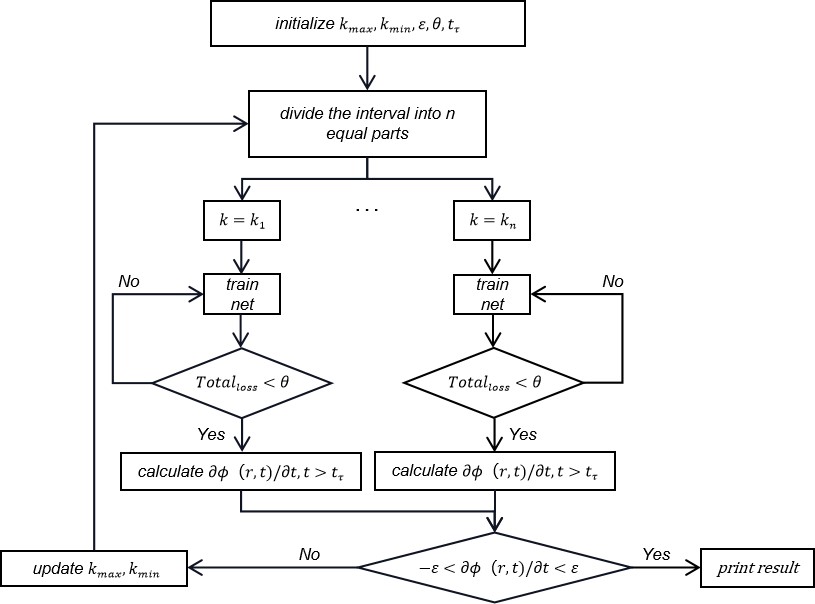}
\caption{Search Flowchart\cite{liu}.}
\label{fig:8}
\end{figure*}

In the absence of delayed neutrons, when the system approaches criticality, regardless of different boundary conditions, the system will converge to a critical state within a very short time (5-8ms), where $\phi$ remains unchanged. At this point, the $\phi$ distribution is the steady-state analytical solution. The interval is partitioned into $n$ equal parts and obtains multiple $k_\infty$ values. The network is trained for each value, calculates $\phi_t$ after a specific time $t_\tau$, and performs a search until $\phi_t$ tends to zero within an acceptable accuracy range. The specific process is illustrated in Fig.~\ref{fig:8}.

In each network, Eq.~\ref{eq:2} is used as the equation for $Loss_{PDE}$. Eq.~\ref{eq:12} is used as the equation for $Loss_{Initial}$ and calculates the mean squared error (MSE) between the predicted values and 0, which serves as the $Loss_{Boundary}$.

Using the automatic differentiation mechanism of NN, the partial derivative of $\phi$ concerning $t$ is calculated after obtaining the predicted values. Given the network's ability to predict flux variations within the time interval of 0.015s, the experiment uses the last five time points to calculate $\partial \phi(r,t)/\partial t$ and determine whether the system is in a critical state \cite{liu}.

\begin{figure}
\centering
\subfigure[$\phi_0=\phi_1$]{
\includegraphics[width=2.6in]{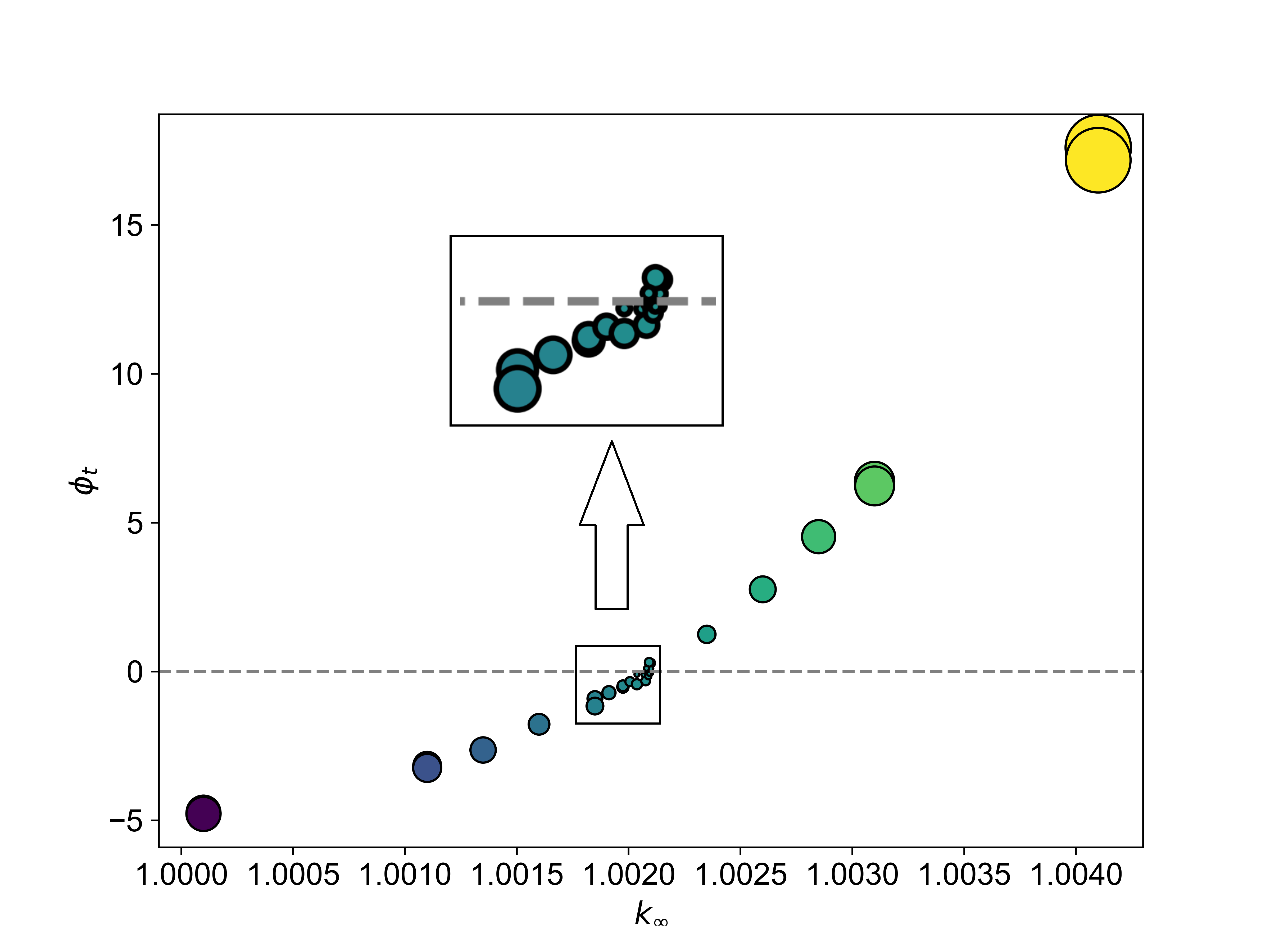}
}
\subfigure[$\phi_0=\phi_2$]{
\includegraphics[width=2.6in]{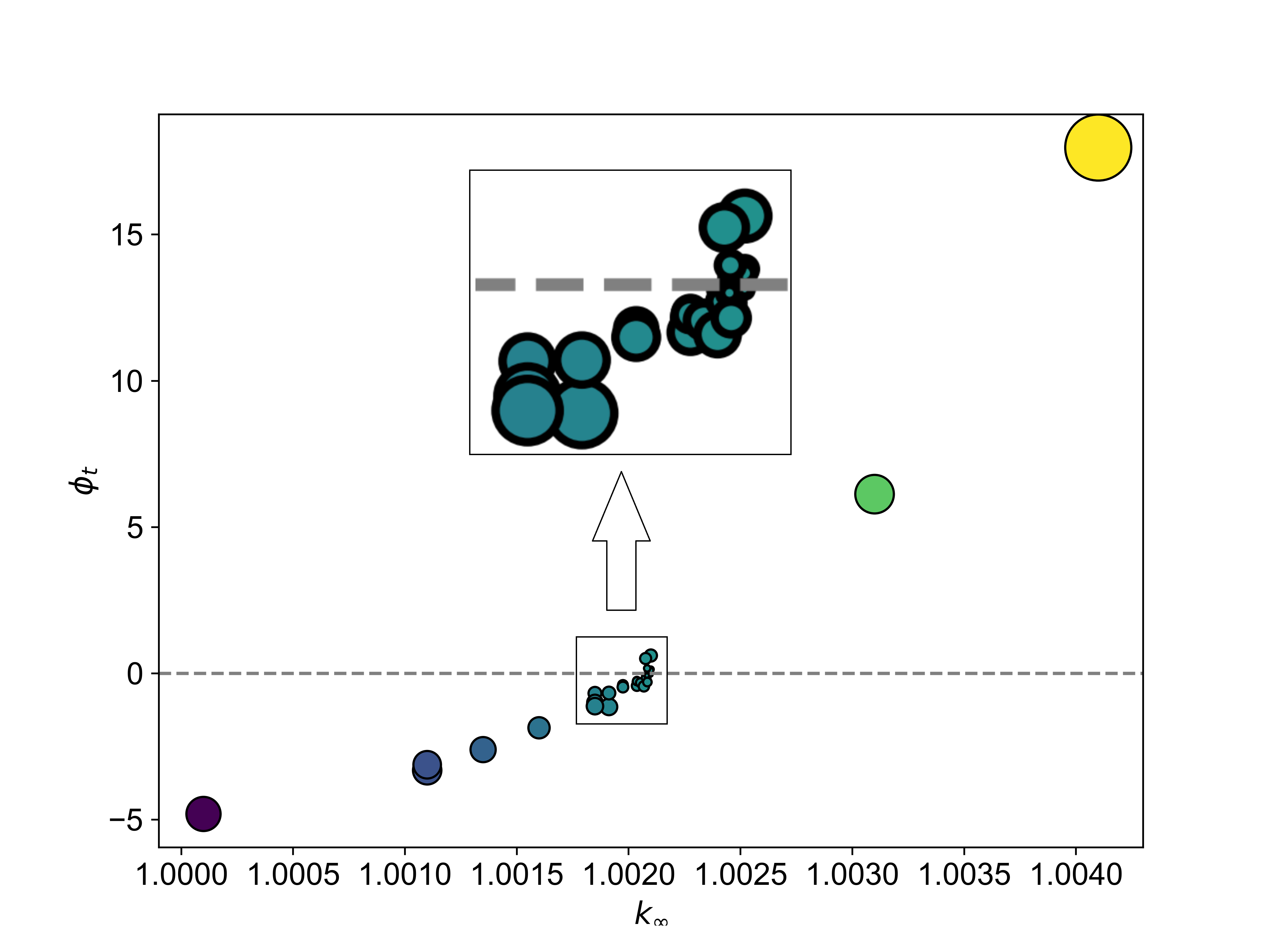}
}
\DeclareGraphicsExtensions.
\caption{Search Result of $\phi_t$.}
\label{fig:9}
\end{figure}

Using Fig.~\ref{fig:8}, the searches for the $k_{eff}$ when the initial distribution follows $\phi_1$ and $\phi_2$ are conducted and record each search result. The $\phi_t$ as a function of $k$ is shown in Fig.~\ref{fig:9}. When $n=2$ and the grid method degenerates into a binary search, it only takes about 20 iterations to obtain the search results with an accuracy level of e-05. The total runtime for the program is approximately 30 minutes.

\begin{figure}[H]
	\centering  
	\subfigbottomskip=1pt 
	\subfigcapskip=-5pt 
	\subfigure[$\phi_0=\phi_1$]{
		\includegraphics[width=0.47\linewidth]{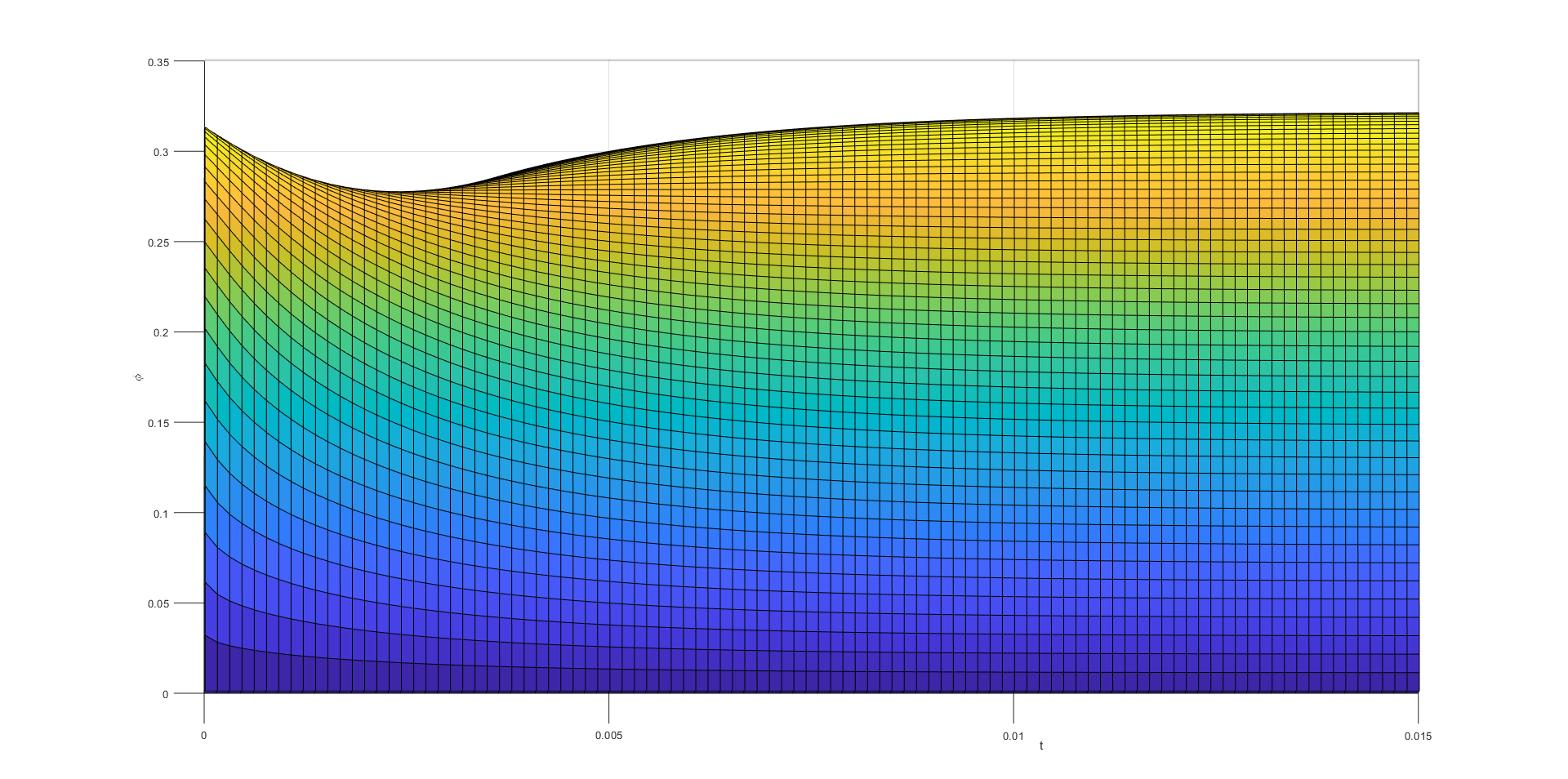} }
	\subfigure[$\phi_0=\phi_2$]{
		\includegraphics[width=0.47\linewidth]{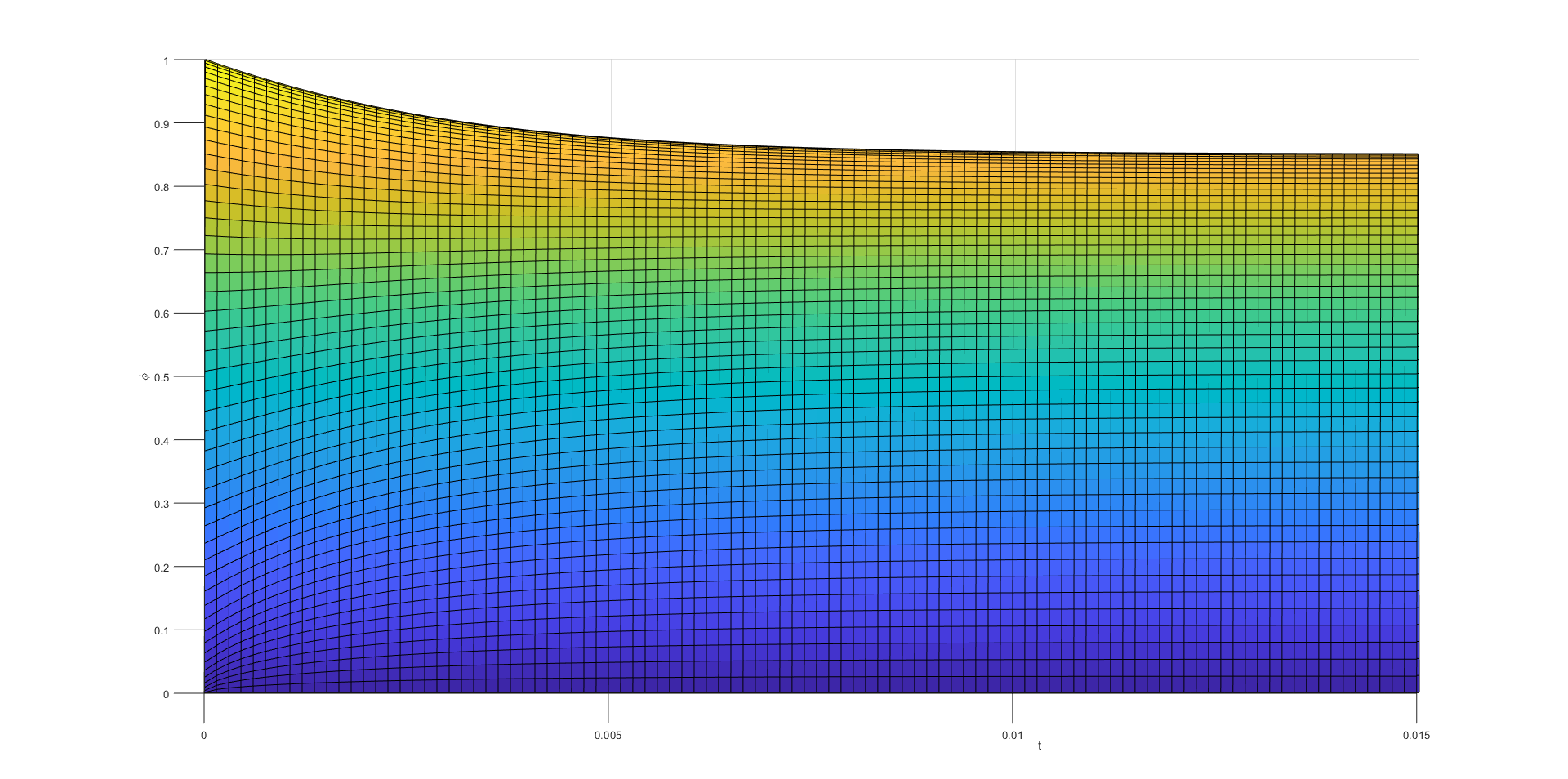} }
	\caption{Steady-State Verification.}
 \label{fig:12}
\end{figure}

From Fig.~\ref{fig:9}, $\phi_t$ changes exponentially with $k_\infty$.  This further suggests that gradient-based methods such as gradient descent or Newton's method can be explored to search for parameter values more quickly and efficiently. Alternatively, using curve fitting techniques to fit the scattered data and the intersection of the resulting curve with the x-axis gives the parameter value $k_\infty$ at the critical state. This may finish within a low number of search iterations. The $k_\infty$ value corresponding to the critical state can be identified by progressively refining the intervals. Search results are shown in Table \ref{tab:4}. Among them, $\Delta \phi$ in the last five time points were recorded to measure whether $\phi$ reaches a steady state. Compared with the result of FCN\cite{liu}, R$^2$-PINN has a smaller $\Delta \phi$, and this means R$^2$-PINN search has higher accuracy than FCN. Further, the search results are validated by generating the $\phi$ distribution for the obtained $k_\infty$ value, which are shown in Fig.~\ref{fig:12}. 

\begin{table}[!htb]
\caption{Search Result.}
\label{tab:4}
\begin{tabular*}{\linewidth}{@{\extracolsep{\fill}}lcccc}
\toprule
& \multicolumn{2}{c}{$\phi_0=\phi_1$} & \multicolumn{2}{c}{$\phi_0=\phi_2$}  \\
\cmidrule(lr){2-3} \cmidrule(lr){4-5} 
 & $k_\infty$ & $\Delta \phi$ & $k_\infty$ & $\Delta \phi$ \\
\midrule
FCN & 1.00202 & -0.017973 & 0.99803 & -0.788704\\ \addlinespace
R$^2$-PINN  & 1.0020822 & -0.001522 & 1.0020824 & -0.001519  \\ \addlinespace
\bottomrule
\end{tabular*}
\end{table}

From Fig.~\ref{fig:12}, it can be observed that the flux tends to stabilize as t increases, indicating that the system reaches a critical state. This indicates that our network can effectively search for the optimal parameters corresponding to the critical state.

\subsection{$k_{eff}$ Search Efficiency Improvement}\label{sec:45}
When starting to search the $k_{eff}$, the initial search interval is large. The k-value to be searched is far away from the critical state k-value. Thus, the accuracy of prediction is not highly required; only use the prediction to compute $\phi_t$ to serve as a priori information for the next interval refinement. So, during the training section, at regular intervals, check if the rate of change of fluxes converges. When $\phi_t$ is converged, stop the training section and soon begin the next k-value network training. This will result in a faster parameter search without loss of accuracy.
The equation for determining when to stop the network iterating is shown below:

\begin{equation}\label{eq:stop}
((\phi_t)_{i+1}-(\phi_t)_i)/((\phi_t)_i-(\phi_t)_{i-1})<\lambda 
\end{equation}

The parameter $\lambda$ is set to 0.01, and every 200 iterations, compute the Eq.~\ref{eq:stop} to determine whether the network stops iterating. After using it, the comparative result when $\phi_0=\phi_1$ is shown in Table \ref{tab:time_consume}. The early termination mechanism can significantly reduce the search time, which is just 0.56 times the original one.

\begin{table}[htb]
\caption{Time Consume Comparation.}
\label{tab:time_consume}
\begin{tabular*}{\linewidth}{@{\extracolsep{\fill}}>{\raggedright\arraybackslash}p{4.5cm}>{\centering\arraybackslash}p{1.5cm}>{\centering\arraybackslash}p{2cm}@{}}
\toprule
& Cost Time & Search Result \\
\midrule
FCN  & 7841s & 1.00202 \\
R$^2$-PINN  & 1837s & 1.00208 \\
R$^2$-PINN(Early Termination)  & 1027s & 1.00207 \\
\bottomrule  
\end{tabular*}
\end{table}

\begin{figure}[!htb]
\includegraphics
  [width=0.6\hsize]
  {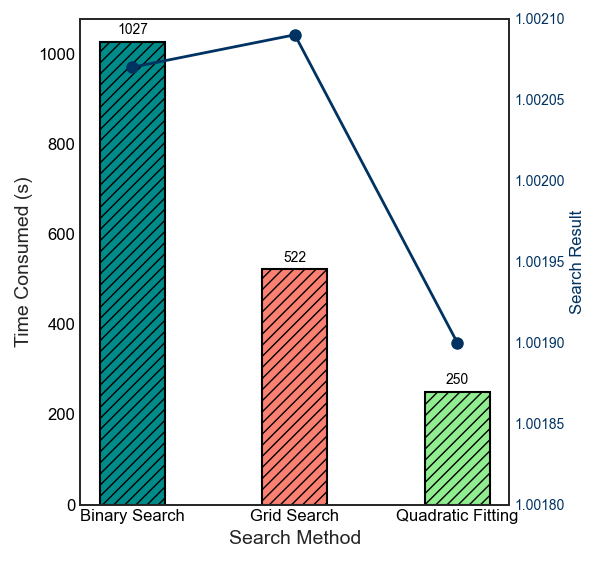}
\caption{Comparison of Search Method.}
\label{fig:time_consume}
\end{figure}

On the other hand, according to Fig.~\ref{fig:9}, it can be seen that $\phi_t$ is exponentially related to $k$. Therefore, it can be used to find the critical corresponding value of $k$ by fitting the quadratic function. To optimize search algorithms, the experiment was conducted to compare three search methods: binary search method, grid search method, and quadratic fitting search method. Using R$^2$-PINN with an early termination mechanism, the results are shown in the following in Fig.~\ref{fig:time_consume}.

The quadratic fitting search method quickly finds the $k_{eff}$ in 250s and reaches e-04 accuracy, while the grid search method can find the $k_{eff}$ in 522s and reach e-05 accuracy. Different methods can be chosen for parameter search based on the actual need for time and accuracy.

\subsection{Network Prediction for Different $k$: R$^2$-PINN's 
Robustness}\label{sec:46}
\begin{figure}[!htb]
\includegraphics
  [width=0.9\hsize]
  {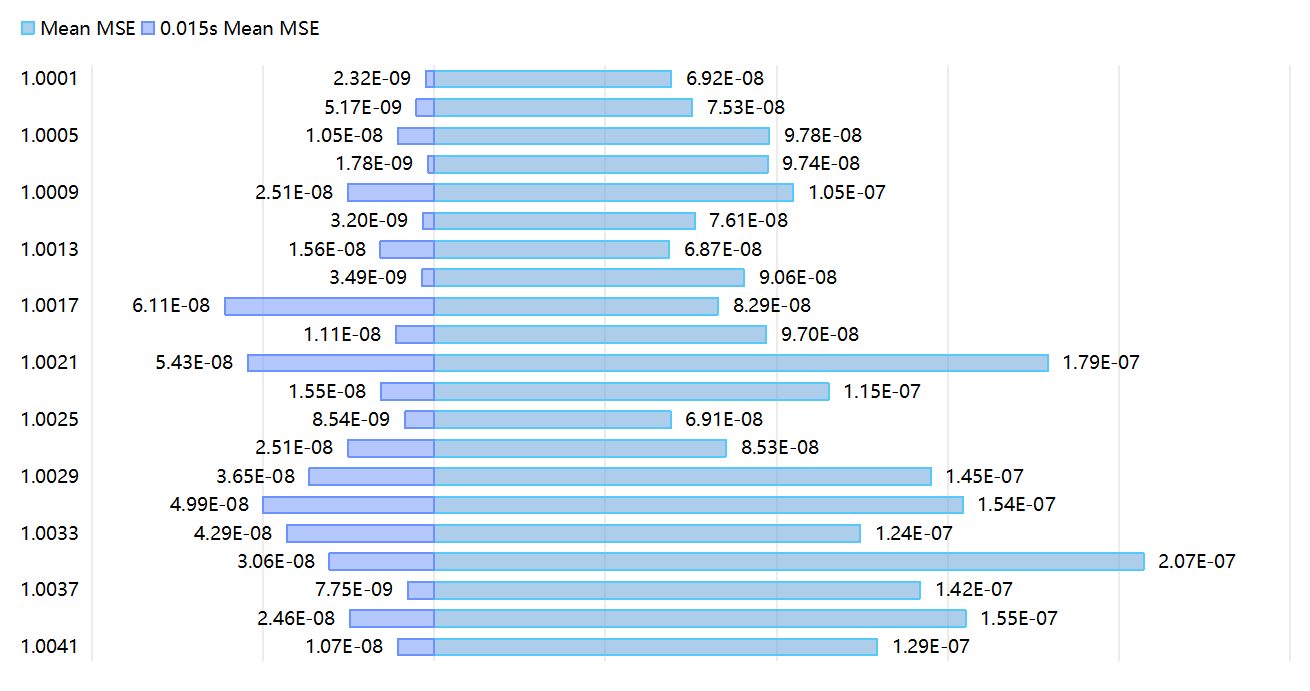}
\caption{MSE Accuracy Comparison in Differenct $k_\infty$.}
\label{fig:14}
\end{figure}

The network is trained at different $k_\infty$ values and compared results with the analytical solution generated to obtain the accuracy at each $k_\infty$ value, as shown in Fig.~\ref{fig:14}, to validate the robustness of R$^2$-PINN.

The robustness of the R$^2$-PINN network is evident from Fig.~\ref{fig:14}. Across all tested values, the network consistently maintains a high accuracy of e-07 or higher, indicating its ability to handle various scenarios and maintain reliable predictions. The observed differences in MSE across different parameter values are relatively small, further highlighting the network's stability and effectiveness. The robustness of the R$^2$-PINN network is evident from its consistently high accuracy across different parameter values. 

\subsection{R$^2$-PINN for solving a Two-dimensional reactor diffusion equation for a single energy group}\label{sec:47}

In this experiment, an 11-layer S-CNN network is used for training, and the rest of the hyperparameters are chosen as in Section \ref{sec:44}.

To validate the generalizability of models, the experiment is conducted between different $k_{\infty}$, and the results are shown in Table \ref{tab:suanli2}, where the network achieves e-5 accuracy under different parameters, and the MSE for the extrapolation region, i.e., the t = 1 region, still achieves e-5 accuracy, which proves that the model performs well in temporal extrapolation ability and can be used to search for the parameters corresponding to the critical state.

\begin{table}[htbp]
  \centering
  \caption{MSE Results in Different $k_{\infty}$.}
  \label{tab:suanli2}
  \begin{tabular*}{0.8\columnwidth}{@{\extracolsep{\fill}}p{1cm} c c@{}}
    \toprule
    $k_{\infty}$ & MSE & t=1s MSE \\
    \midrule
    1.1 & 4.4e-6 & 9.0e-6 \\
    1.11 & 1.1e-6 & 1.7e-6 \\
    1.12 & 3.2e-6 & 5.2e-6 \\
    1.13 & 1.2e-6 & 2.2e-6 \\
    1.14 & 1.8e-6 & 3.0e-6 \\
    1.15 & 8.5e-7 & 1.6e-6 \\
    1.16 & 2.4e-6 & 4.4e-6 \\
    1.17 & 1.1e-5 & 2.2e-5 \\
    1.18 & 5.5e-6 & 9.3e-6 \\
    1.19 & 2.2e-6 & 4.7e-6 \\
    1.2 & 4.5e-6 & 8.2e-6 \\
    \bottomrule
  \end{tabular*}
\end{table}

\begin{figure}[H]
	\centering  
	\subfigbottomskip=1pt 
	\subfigcapskip=-5pt 
	\subfigure[Predict Result(t=0s).]{
		\includegraphics[width=0.48\linewidth]{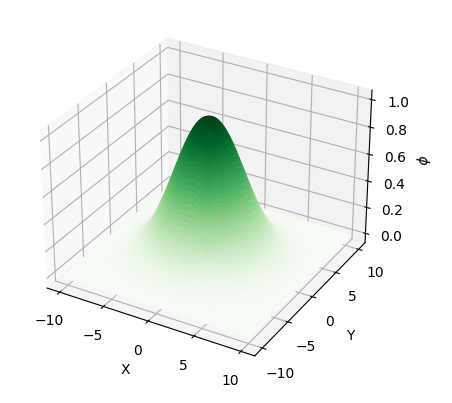}}
	\subfigure[Predict Result(t=1s).]{
		\includegraphics[width=0.48\linewidth]{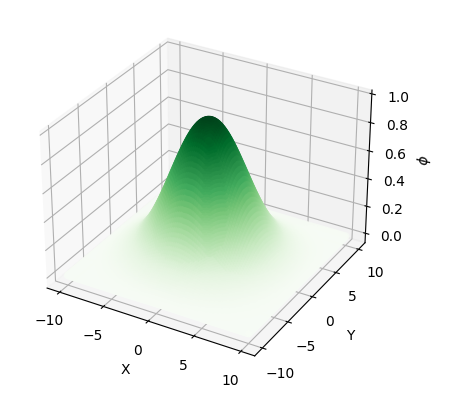}}
	  \\
	\subfigure[Truth(t=0s).]{
		\includegraphics[width=0.48\linewidth]{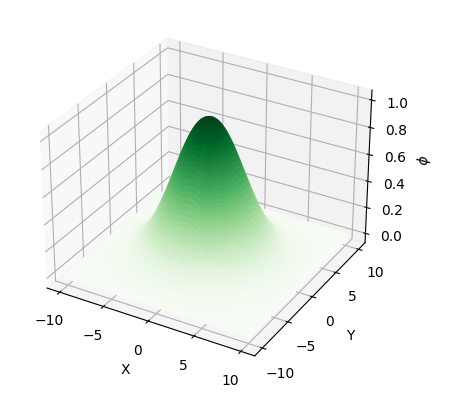}}
	\subfigure[Truth(t=1s).]{
		\includegraphics[width=0.48\linewidth]{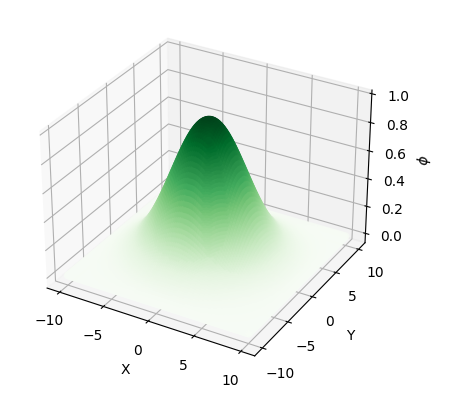}}
	\caption{Comparison Between Result and Truth.}
\label{fig:suanli2}
\end{figure}

\begin{figure}[H]
	\centering  
	\subfigbottomskip=1pt 
	\subfigcapskip=-5pt 
	\subfigure[t=0s]{
		\includegraphics[width=0.47\linewidth]{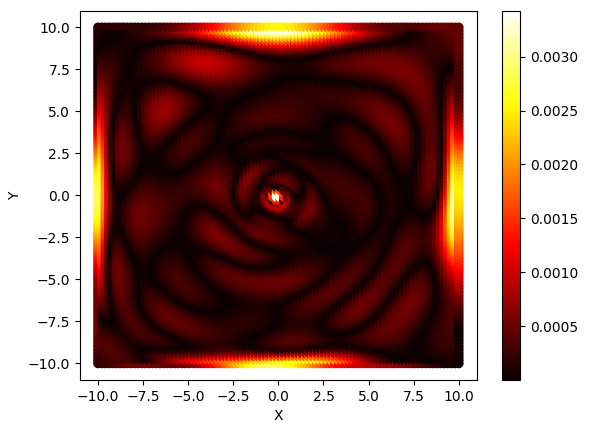} }
	\subfigure[t=1s]{
		\includegraphics[width=0.47\linewidth]{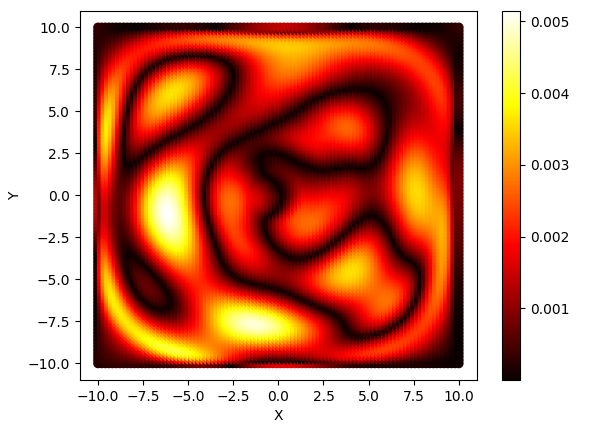} }
	\caption{Loss Field.}
 \label{fig:suanli2_loss}
\end{figure}

Searching for the $k_{\infty}$ parameter in the same way as in Section \ref{sec:44} yields $k_{eff}$ = 1.1378. While using the source iteration method find $k_{\infty}$= 1.1202. The calculation error is around 1.6\%. The results of the prediction and the reference truth are shown in Fig.~\ref{fig:suanli2}. The error plot is shown in Fig.~\ref{fig:suanli2_loss}.

\subsection{R$^2$-PINN for solving Two-dimensional rectangular geometry multigroup multi-material diffusion problem}\label{sec:48}
\begin{figure}[H]
	\centering  
	\subfigbottomskip=2pt 
	\subfigcapskip=-5pt 
	\subfigure[$\phi_1$ Distribution in x-z Plane]{
		\includegraphics[width=0.45\linewidth]{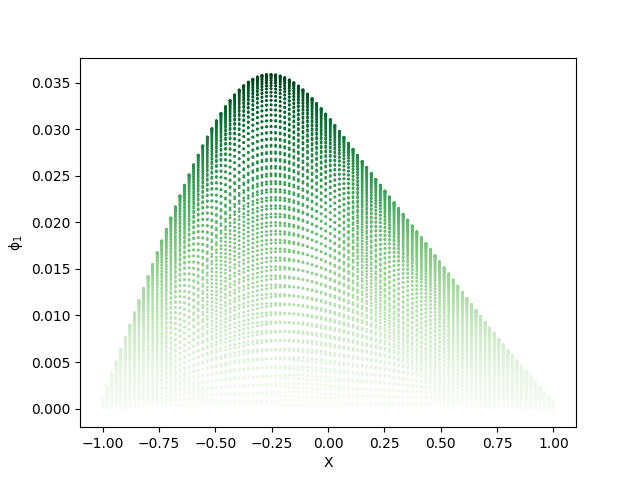}}
	\subfigure[$\phi_2$ Distribution in x-z Plane]{
		\includegraphics[width=0.45\linewidth]{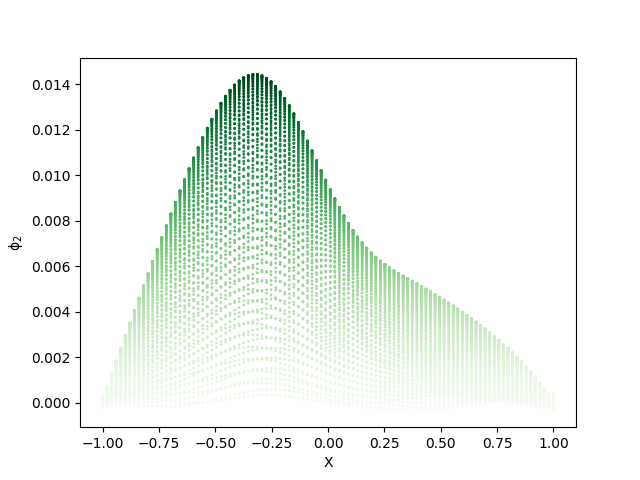}}
	  \\
	\subfigure[$\phi_1$ Distribution in y-z Plane]{
		\includegraphics[width=0.45\linewidth]{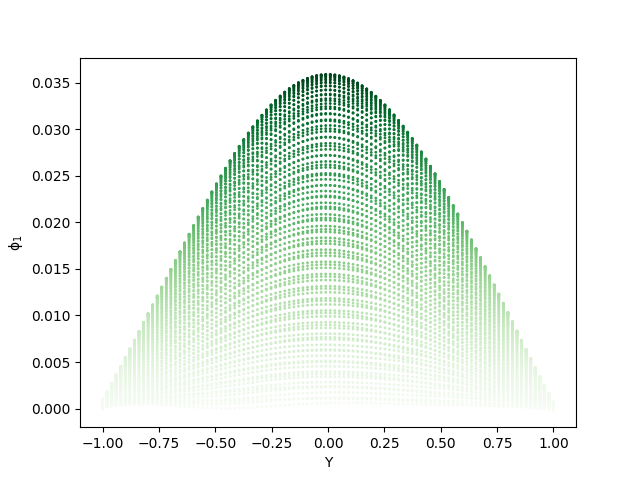}}
	\subfigure[$\phi_2$ Distribution in y-z Plane]{
		\includegraphics[width=0.45\linewidth]{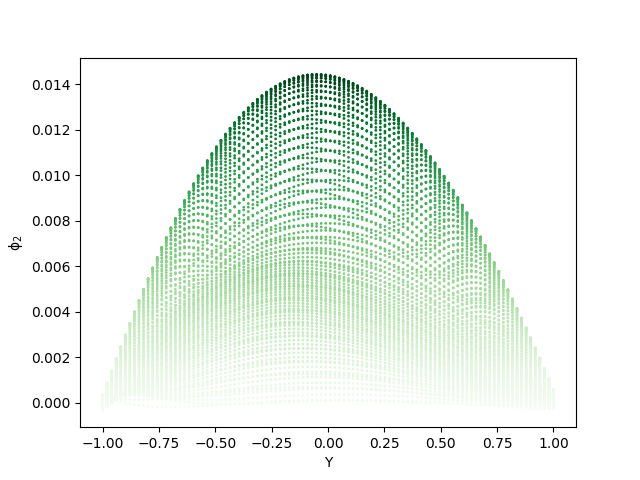}}
	\caption{Predict Result.}
 \label{fig:suanli3_predict}
\end{figure}

\begin{figure}[H]
	\centering  
	\subfigbottomskip=2pt 
	\subfigcapskip=-5pt 
	\subfigure[$\phi_1$ Distribution in x-z Plane]{
		\includegraphics[width=0.45\linewidth]{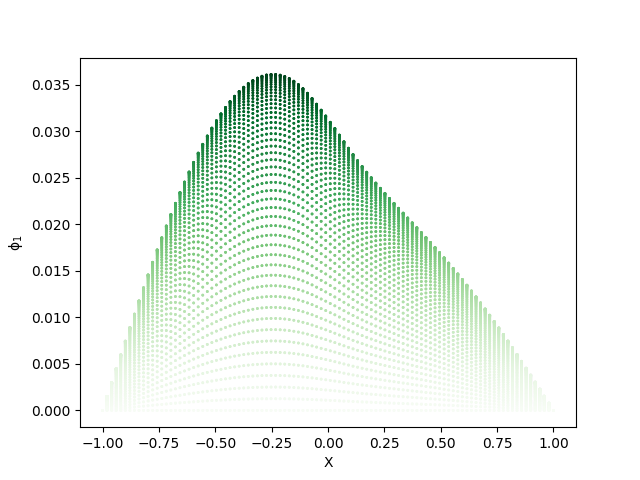}}
	\subfigure[$\phi_2$ Distribution in x-z Plane]{
		\includegraphics[width=0.45\linewidth]{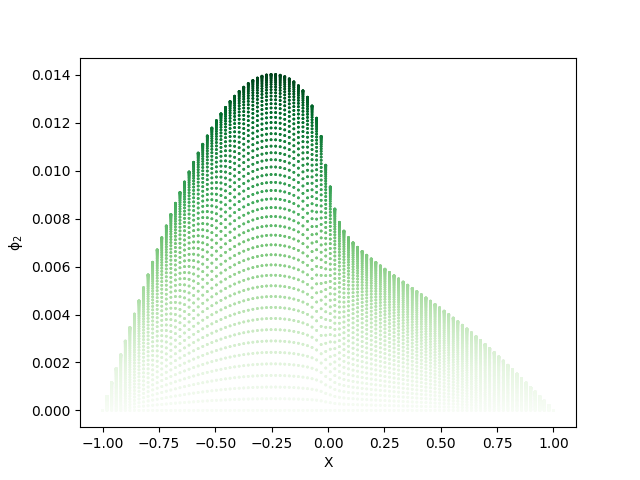}}
	  \\
	\subfigure[$\phi_1$ Distribution in y-z Plane]{
		\includegraphics[width=0.45\linewidth]{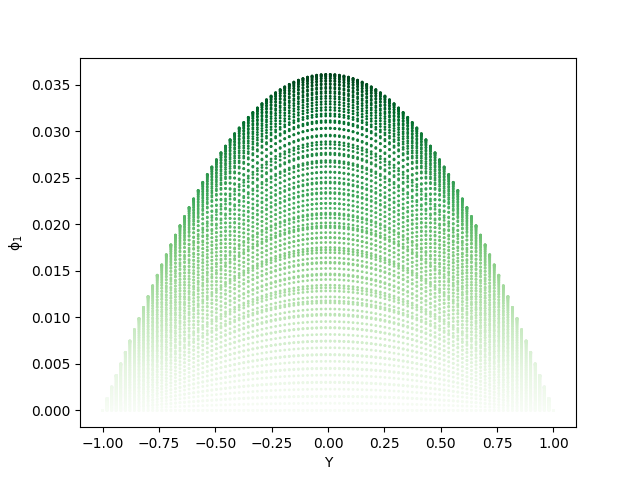}}
	\subfigure[$\phi_2$ Distribution in y-z Plane]{
		\includegraphics[width=0.45\linewidth]{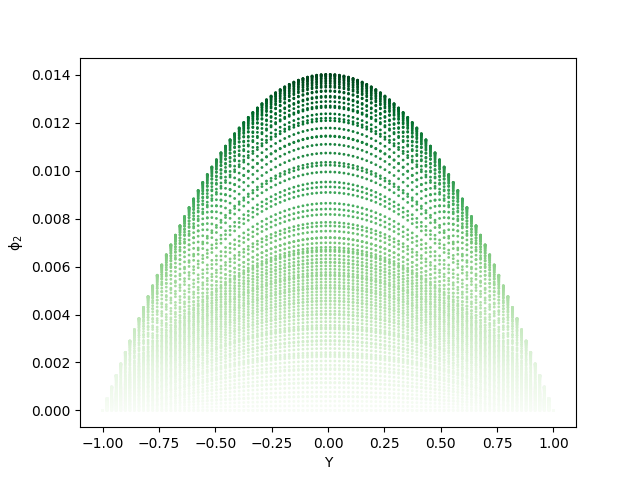}}
	\caption{Truth Result.}
 \label{fig:suanli3_truth}
\end{figure}

In this two-group problem, we separately predict the neutron flux of the fast group ($\phi_1$) and the thermal group ($\phi_2$) using dual PINNs at ( $k_{eff}=0.9693$ ). As there is an interconversion relationship between the fluxes, the dual PINNs share losses and undergo sequential iterative optimization. The experiment utilizes a 4-layer S-CNN for training. Other PINN parameters remain consistent with those in Section \ref{sec:44}.
Using R$^2$-PINN, MSE of $\phi_1$ reaches 1.36e-06, MSE of $\phi_2$ reaches 2.49e-07. The results and reference solutions are shown in Fig.~\ref{fig:suanli3_predict}, Fig.~\ref{fig:suanli3_truth}.

\begin{figure}[H]
	\centering  
	\subfigbottomskip=1pt 
	\subfigcapskip=-5pt 
	\subfigure[$\phi_1$ Loss]{
		\includegraphics[width=0.47\linewidth]{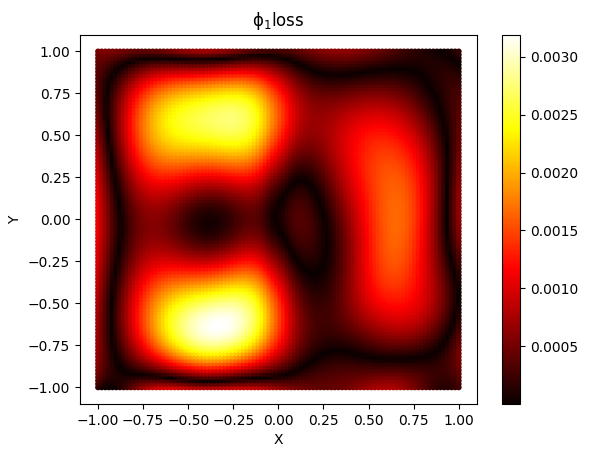} }
	\subfigure[$\phi_2$ Loss]{
		\includegraphics[width=0.47\linewidth]{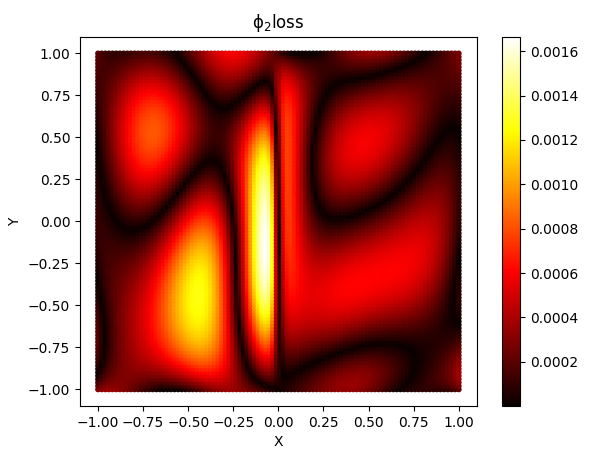} }
  \\
	\subfigure[$\phi_1$ Loss Rate]{
		\includegraphics[width=0.47\linewidth]{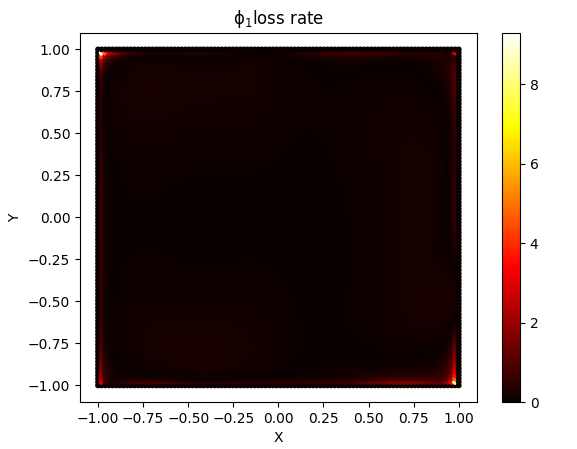}}
	\subfigure[$\phi_2$ Loss Rate]{
		\includegraphics[width=0.47\linewidth]{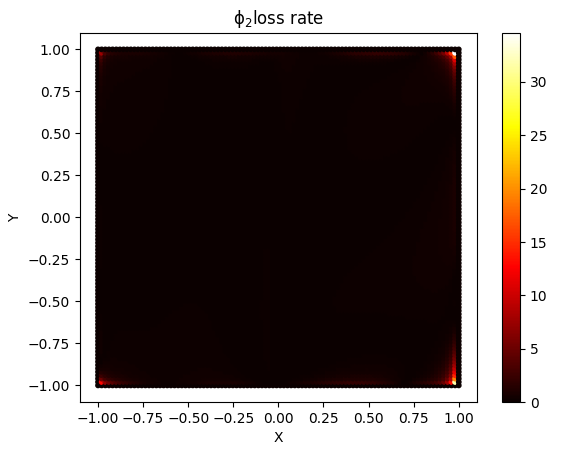}}
	\caption{Loss Field and Loss Rate Field.}
 \label{fig:suanli3_loss}
\end{figure}

 Due to the small value of the flux, to make the error representation clearer, we also calculated the loss rate and displayed it in Fig.~\ref{fig:suanli3_loss}. Specific equation are shown in Eq.~\ref{eq:lossrate}.
 
 \begin{equation}\label{eq:lossrate}
loss\ rate=ABS(\phi_{predict}-\phi_{truth})/\phi_{truth}
\end{equation}

The results show that we can effectively find the steady-state flux distribution under the specified $k_{eff}$ according to the steady-state equations, and at the same time, according to the experiments in Section \ref{sec:44} and Section \ref{sec:47}, we are also able to find the critical parameters according to the transient equations, so the model can be well adapted to and solved the two major problems of nuclear reactors, i.e., solving for the steady-state fluxes as well as solving for the steady-state parameters.

\subsection{R$^2$-PINN for solving 2D-IAEA problem}\label{sec:49}
Using a 6-layer R$^2$-PINN structure, incorporates $k_{eff}$ as a parameter for iterative optimization in the neural network training. The model utilized 18000 points for compute $Loss_{PDE}$, 500 points for $Loss_{Boundary}$ and 76 points for $Loss_{data}$. 

\begin{figure}[H]
	\centering  
	\subfigbottomskip=1pt 
	\subfigcapskip=-5pt 
	\subfigure[R$^2$-PINN result of $\phi_1$]{
		\includegraphics[width=0.47\linewidth]{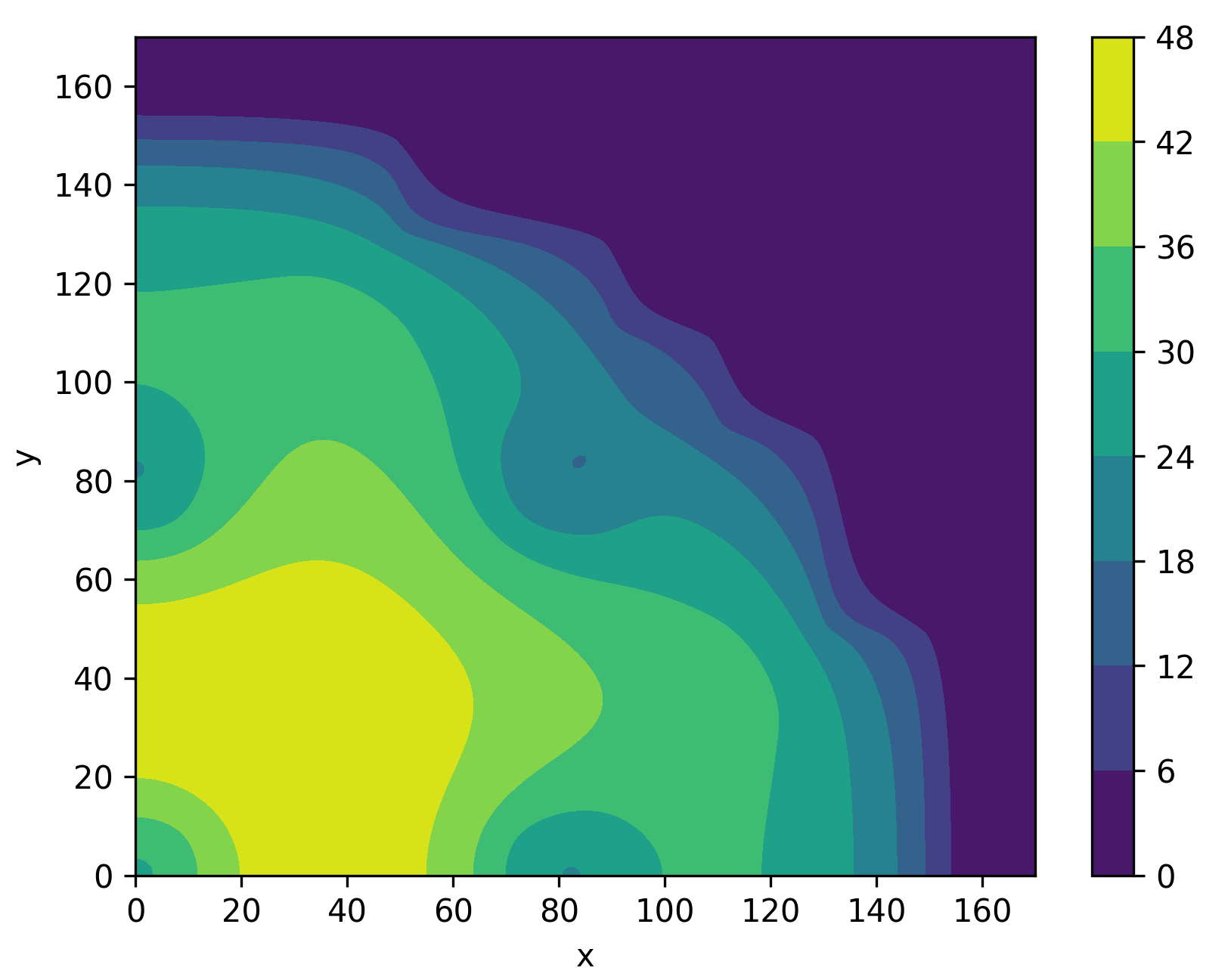} }
	\subfigure[R$^2$-PINN result of $\phi_2$]{
		\includegraphics[width=0.47\linewidth]{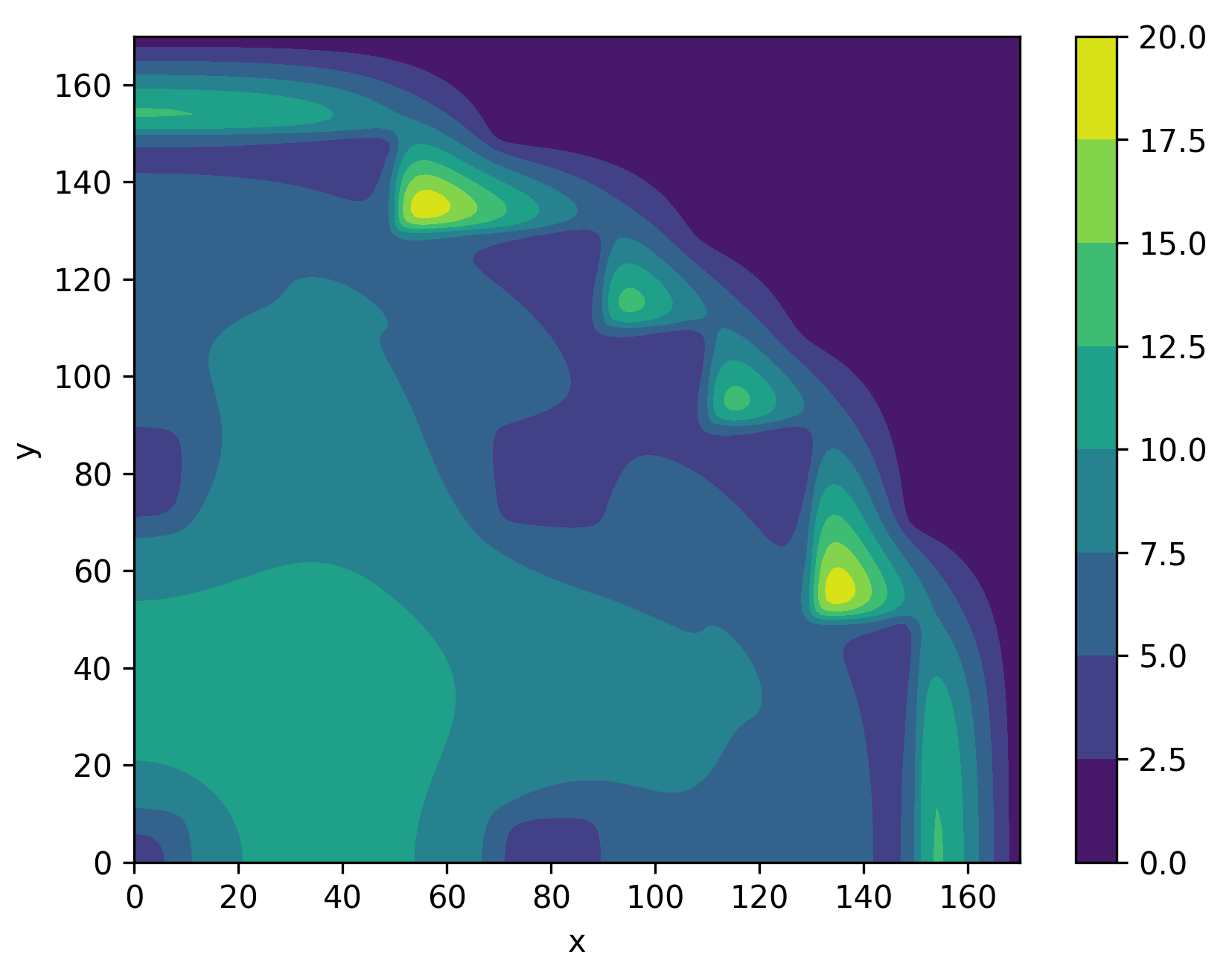} }
  \\
	\subfigure[FreeFem++ result of $\phi_1$]{
		\includegraphics[width=0.47\linewidth]{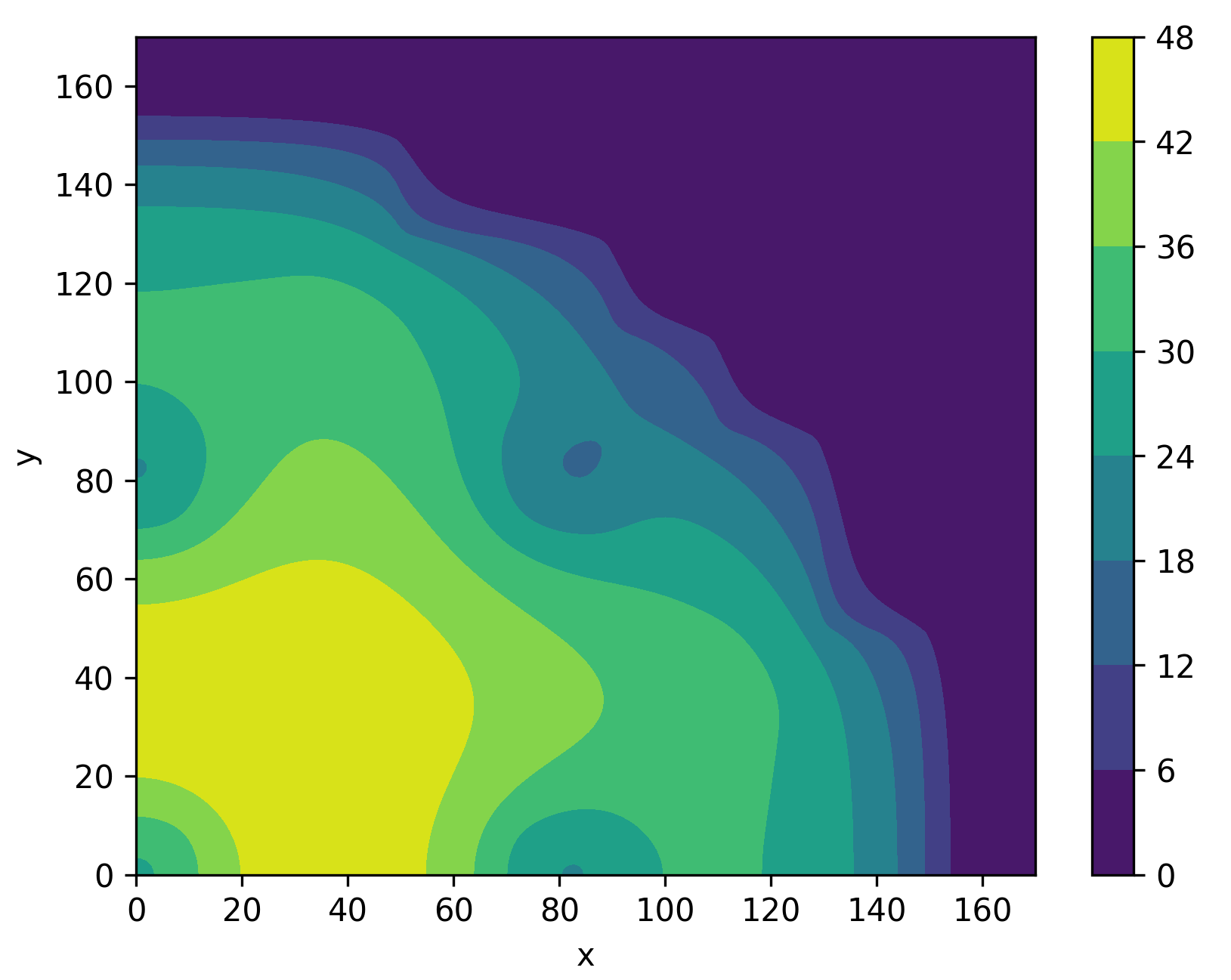}}
	\subfigure[FreeFem++ result of $\phi_2$]{
		\includegraphics[width=0.47\linewidth]{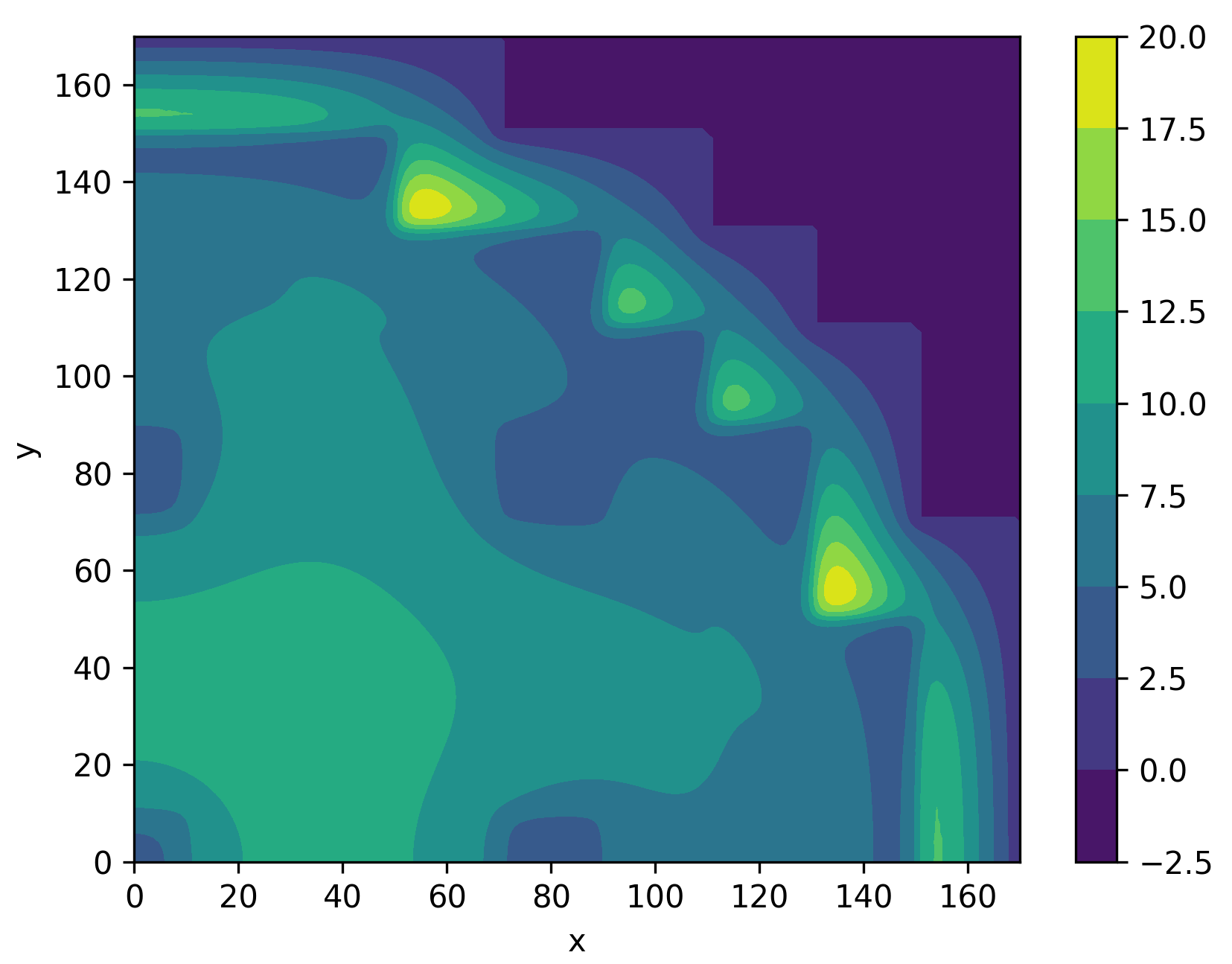}}
	\caption{Comparison Between Results and Reference.}
 \label{fig:suanli4_predict}
\end{figure}

\begin{figure}[H]
	\centering  
	\subfigbottomskip=1pt 
	\subfigcapskip=-5pt 
	\subfigure[Absolute error of $\phi_1$]{
		\includegraphics[width=0.47\linewidth]{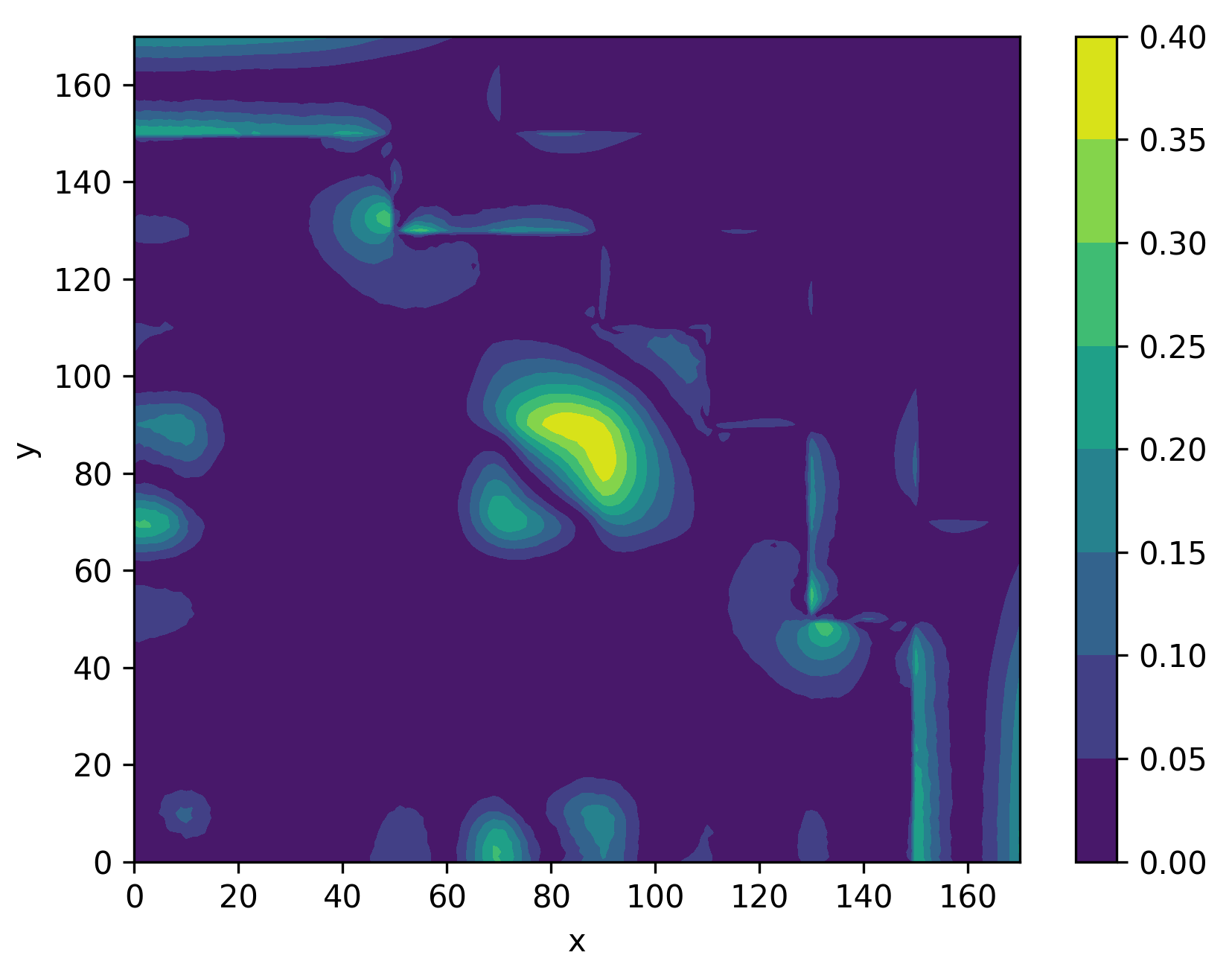} }
	\subfigure[Absolute error of $\phi_2$]{
		\includegraphics[width=0.47\linewidth]{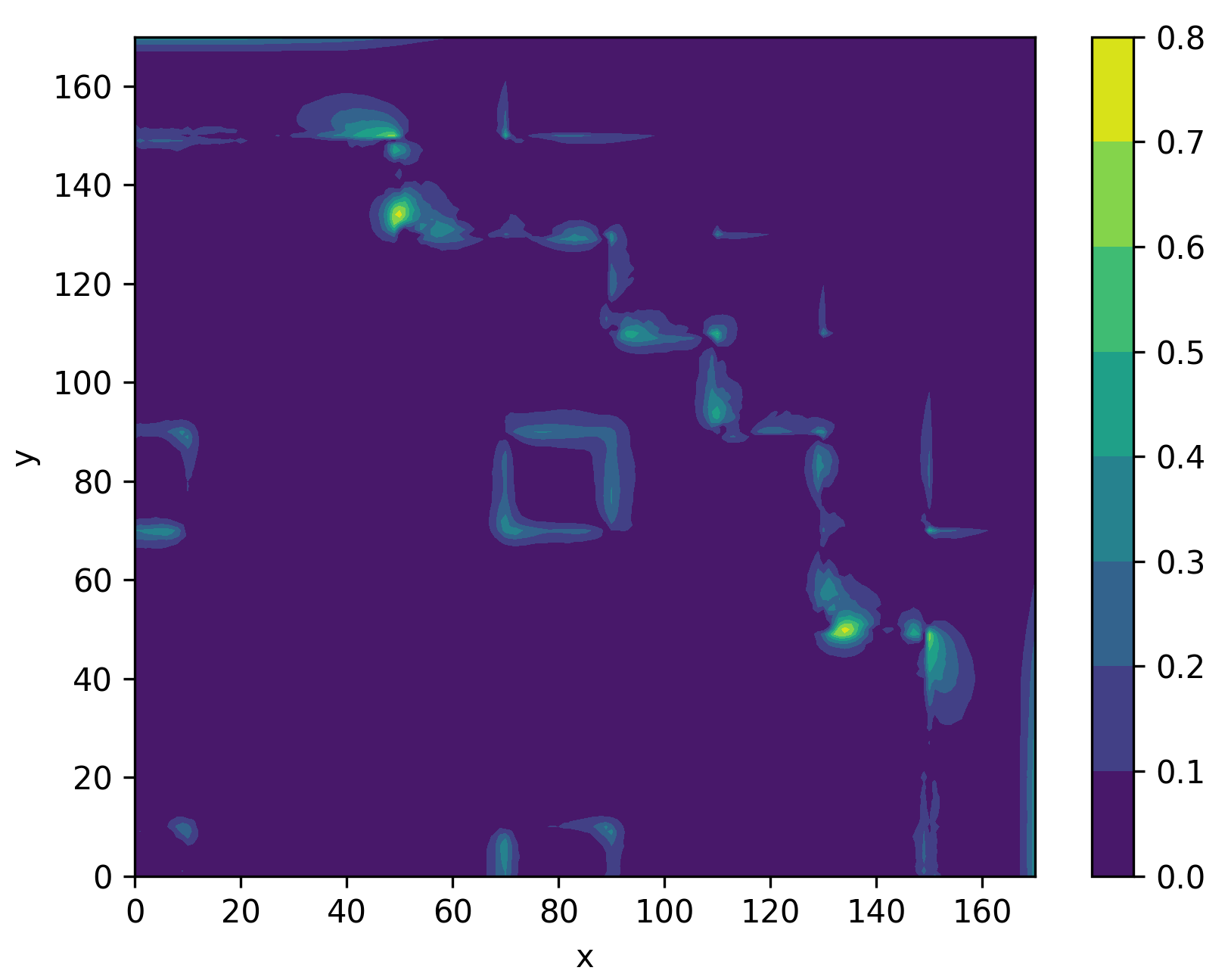} }
	\caption{Absolute Loss Field.}
 \label{fig:suanli4_loss}
\end{figure}

Ultimately, relative error and relative $L_{\infty}$ error are used to evaluate the accuracy of R$^2$-PINN, with relative $L_{\infty}$ error being particularly important in the nuclear engineering domain. The specific equations are shown in Eq.~\ref{eq:er} and Eq.~\ref{eq:el}. The prediction results and the reference truth are presented in Fig.~\ref{fig:suanli4_predict}. Additionally, the absolute error plot is depicted in Fig.~\ref{fig:suanli4_loss}.

\begin{equation}\label{eq:er}
e_r=ABS(\phi_{predict}-\phi_{truth})/\phi_{truth}
\end{equation}
\begin{equation}\label{eq:el}
e_{\infty}=\frac{\left \| \phi_{predict} -\phi_{truth} \right \| _{\infty }}{\left \| \phi_{truth} \right \| _{\infty }} 
\end{equation}
\begin{table}[htbp]
  \centering
  \caption{Results of 2D-IAEA benchmark.}
  \label{tab:suanli4}
  \begin{tabular}{cccc} 
    \toprule
    $k_{eff}$ & $e_r$ of $k_{eff}$ & $e_{\infty}$ of $\phi_1$ & $e_{\infty}$ of $\phi_2$\\
    \midrule
    1.02977 & 1.797e-4 & 0.008 & 0.038 \\
    \bottomrule
  \end{tabular}
\end{table}

Considering the engineering acceptance criteria for the 2D IBP, the flux calculation error in fuel assemblies with a relative flux higher than 0.9 must be less than 5\%, and in fuel assemblies with a relative flux less than 0.9, the flux calculation error must be less than 8\%. Additionally, the relative error of $k_{eff}$ must be less than 0.005 \cite{dataPINN}. The predicted results meet these acceptance criteria, indicating that R2-PINN also has practical engineering application value.

\section{Results and Discussion}\label{sec:RD}
The ablation study on the number of layers revealed that increasing the depth of the S-CNN architecture improved accuracy. Notably, a kernel size of 1 yielded superior results. However, there was an accuracy bottleneck, making overly deep layers unnecessary. In Section \ref{sec:42}, S-CNN outperformed the FCN architecture across various layer configurations. Even when the number of layers was doubled, vanishing gradients did not occur. Adding skip connections effectively mitigated vanishing gradients, enhancing the model's stability, robustness, and overall accuracy.

Meanwhile, in Section \ref{sec:43}, the sensitivity analysis of the resample parameter revealed that the optimal R$^2$-PINN configuration is a resample granularity of 2 and 500 resamples. Setting the loss weight coefficient $w$ to 100 achieved the best prediction performance, resulting in lower other losses compared to the PDE loss. The test loss in Fig.~\ref{fig:Test} shows a smooth training process without significant fluctuations. Around 2000 epochs, the LBFGS optimizer automatically stops training, achieving an MSE accuracy of e-07, indicating high precision.

Further, $k_{eff}$ is found by searching the critic state in Section \ref{sec:44} using the adjusted R$^2$-PINN. R$^2$-PINN quickly converged in just 1000 epochs and reached the e-08 accuracy. The fast convergence speed and high convergence accuracy of the model suggest that the model is suitable for parameter search goals that require the training of multiple networks.

Then, multiple search methods are compared in Section \ref{sec:45} to improve the search efficiency. From the experiment result, the quadratic fitting search method can quickly find the $k_{eff}$ only in two k-value search processes, with e-04 accuracy in just 250s. This shows the great value in the scenarios with high real-time requirements. Meanwhile, the grid method can also achieve the parameter search with e-05 accuracy within 10 minutes. When high accuracy is required, the grid method can set the optimal grid refine numbers to reach an acceptable duration with higher accuracy.

By conducting experiments in different $k$ in Section \ref{sec:46}, the result shows that our R$^2$-PINN network has exceptional performance in capturing the system's dynamics within the domain $\Omega_1$. This region exhibits a significant improvement in accuracy compared to FCN networks. By accurately representing this specific region's intricate features and sharp gradients, our network enables more precise determinations of whether the system is in a steady state. This enhanced accuracy proves invaluable when conducting parameter searches, allowing for higher precision and more reliable results.

Finally, to verify the generalizability of the model, the experiments were computed using the S-CNN architecture for 2D single-cluster in Section \ref{sec:47} and the parameter search error is around 1.6\%. Using S-CNN solving 2D multi-cluster multi-material in Section \ref{sec:48} gives the solution with e-06 accuracy. Standardized test problem sets——2D-IAEA benchmark for search $k_{eff}$ reach e-4 accuracy in Section \ref{sec:49}. These result shows the model can effectively predict the variation of physical quantities in the physical field under multiple equations, different initial conditions, and boundaries and has good generalization under different scenarios.

In summary, the R$^2$-PINN network performs better than FCN networks in solving neutron diffusion equations. The accuracy improvement of 1-2 points is noteworthy, considering the computational efficiency achieved by our framework. This efficiency reduces the computational burden and maintains high accuracy, making it a promising approach for practical applications. By using a suitable search method, our architecture has good real-time performance.

\section{Conclusions}\label{sec:5}
This paper introduces a novel and innovative framework called R$^2$-PINN that addresses the persistent challenge of the disappearing gradient phenomenon in DNN. Additionally, our framework is designed to enhance the accuracy and computational efficiency of PINNs when solving neutron diffusion equations. R$^2$-PINN can find $k_{eff}$ with e-04 accuracy in just 250s. The single NN accuracy reached e-08 accuracy on average, which is an order of magnitude higher than FCN. The S-CNN architecture is integrated into the R$^2$-PINN framework to overcome the vanishing gradients. By leveraging cross-layer connections, our model effectively learns residual information, thereby improving the depth and expressive power of the network. This architectural enhancement ensures that the network can effectively propagate gradients through the layers, enabling more accurate and stable learning.
Furthermore, the RAR method is introduced to enhance the representation and sampling strategy within the network. The RAR method allows for the adaptive collection of sample points, ensuring that the model captures essential features and gradients in the solution space. This refinement strategy dramatically improves the network's capacity to handle sharp gradients and intricate features in PDE solutions.

In the experimental section, comprehensive comparative experiments were conducted to optimize our R$^2$-PINN framework. Through meticulous parameter tuning, including adjusting the number of layers, kernel size, and resampling hyperparameters, R$^2$-PINN achieved significant accuracy improvements of 1-2 units compared to FCN, demonstrating its effectiveness in enhancing PDE solutions. The parameter search capability of R$^2$-PINN was validated by efficiently determining the corresponding value of $k$ when it enters the critical state within the specified search interval through high-precision network predictions. To evaluate model robustness and accuracy under varying parameters, MSE validation experiments with different values of $k$ were conducted, consistently yielding high accuracy results. Finally, for complex models like the two-dimensional single-group neutron diffusion equation, two-group two-material neutron diffusion equation models, and 2D-IAEA benchmark, the search for effective value-added coefficients and the steady-state flux distribution solution was successfully carried out.

Overall, the experimental results confirm that the integration of S-CNN and the RAR mechanism in R$^2$-PINN empowers the network to capture intricate features and sharp gradients in PDE solutions. The adaptive refinement strategy significantly improves the distribution of residual points, enhancing the network's ability to represent complex physical systems accurately. Our findings provide compelling evidence of the potential of R$^2$-PINN to advance the field of deep learning-based PDE solving. Not only does our framework outperform existing methods, but it also exhibits promising potential for application in real-world physical systems. The contributions of our work hold substantial implications for the development of more accurate and efficient models in various science and engineering domains.

\end{document}